\def\cvprPaperID{6584}
\def\confName{CVPR}
\def\confYear{2023}

\def\paperTitle{{SadTalker: Learning Realistic 3D Motion Coefficients for Stylized Audio-Driven Single Image Talking Face Animation}}

\def\authorBlock{
    Wenxuan Zhang\thanks{Equal contribution} \textsuperscript{~,~\rm 1} \qquad
    Xiaodong Cun\footnotemark[1] \textsuperscript{~,~\rm 2} \qquad
    Xuan Wang\textsuperscript{\rm 3}  \qquad
    Yong Zhang\textsuperscript{\rm 2} \qquad
    Xi Shen\textsuperscript{\rm 2} \\
    Yu Guo\textsuperscript{\rm 1} \qquad
    Ying Shan\textsuperscript{\rm 2} \qquad
    Fei Wang\thanks{Corresponding Author} \textsuperscript{~,~\rm 1 }
    \\
    \\
    \textsuperscript{\rm 1}~Xi'an Jiaotong University \qquad \textsuperscript{\rm 2}~Tencent AI Lab \qquad \textsuperscript{\rm 3}~Ant Group \\
    \\
    \url{https://sadtalker.github.io}
}

\newif\ifreview 
\newif\ifarxiv \newcommand{\arxiv}{\arxivtrue}
\newif\ifcamera 
\newif\ifrebuttal 

\arxiv

\pdfoutput=1
\documentclass[10pt,twocolumn,letterpaper]{article}
\ifreview \usepackage[review]{cvpr} \fi
\ifarxiv \usepackage[pagenumbers]{cvpr} \fi
\ifrebuttal \usepackage[rebuttal]{cvpr} \fi
\ifcamera \usepackage{cvpr} \fi

\usepackage{graphicx}
\usepackage{amsmath}
\usepackage{amssymb}
\usepackage{booktabs}

\usepackage{times}
\usepackage{microtype}
\usepackage{epsfig}
\usepackage[table,xcdraw]{xcolor}
\usepackage{caption}
\usepackage{float}
\usepackage{placeins}
\usepackage{color, colortbl}
\usepackage{stfloats}
\usepackage{enumitem}
\usepackage{tabularx}
\usepackage{xstring}
\usepackage{multirow}
\usepackage{xspace}
\usepackage{url}
\usepackage{subcaption}
\usepackage{xcolor}
\usepackage[hang,flushmargin]{footmisc}
\ifcamera \usepackage[accsupp]{axessibility} \fi

% ...

\ifarxiv  \fi

\newcommand{\R}[1]{{%
    \textbf{%
        \ifstrequal{#1}{1}{\textcolor{red}{R#1}}{%
        \ifstrequal{#1}{2}{\textcolor{blue}{R#1}}{%
        \ifstrequal{#1}{3}{\textcolor{magenta}{R#1}}{%
        \ifstrequal{#1}{4}{\textcolor{teal}{R#1}}{%
                           \textcolor{cyan}{R#1}%
        }}}}%
    }%
}}

\usepackage{xr-hyper}

\makeatletter
\newcommand*{\addFileDependency}[1]{
  \typeout{(#1)}
  \@addtofilelist{#1}
  \IfFileExists{#1}{}{\typeout{No file #1.}}
}

\makeatother

\usepackage[pagebackref,breaklinks,colorlinks]{hyperref}
\usepackage[capitalize]{cleveref}
\crefname{section}{Sec.}{Secs.}
\crefname{table}{Table}{Tables}
\crefname{figure}{Fig.}{Figs.}

\hypersetup{
    colorlinks = true,
    linkbordercolor = {white},
    citecolor=blue
}

\begin{document}
\title{\paperTitle}
\author{\authorBlock}

\twocolumn[{
\maketitle
\begin{center}
    \captionsetup{type=figure}
    \vspace{-2em}
    \includegraphics[width=1.\textwidth]{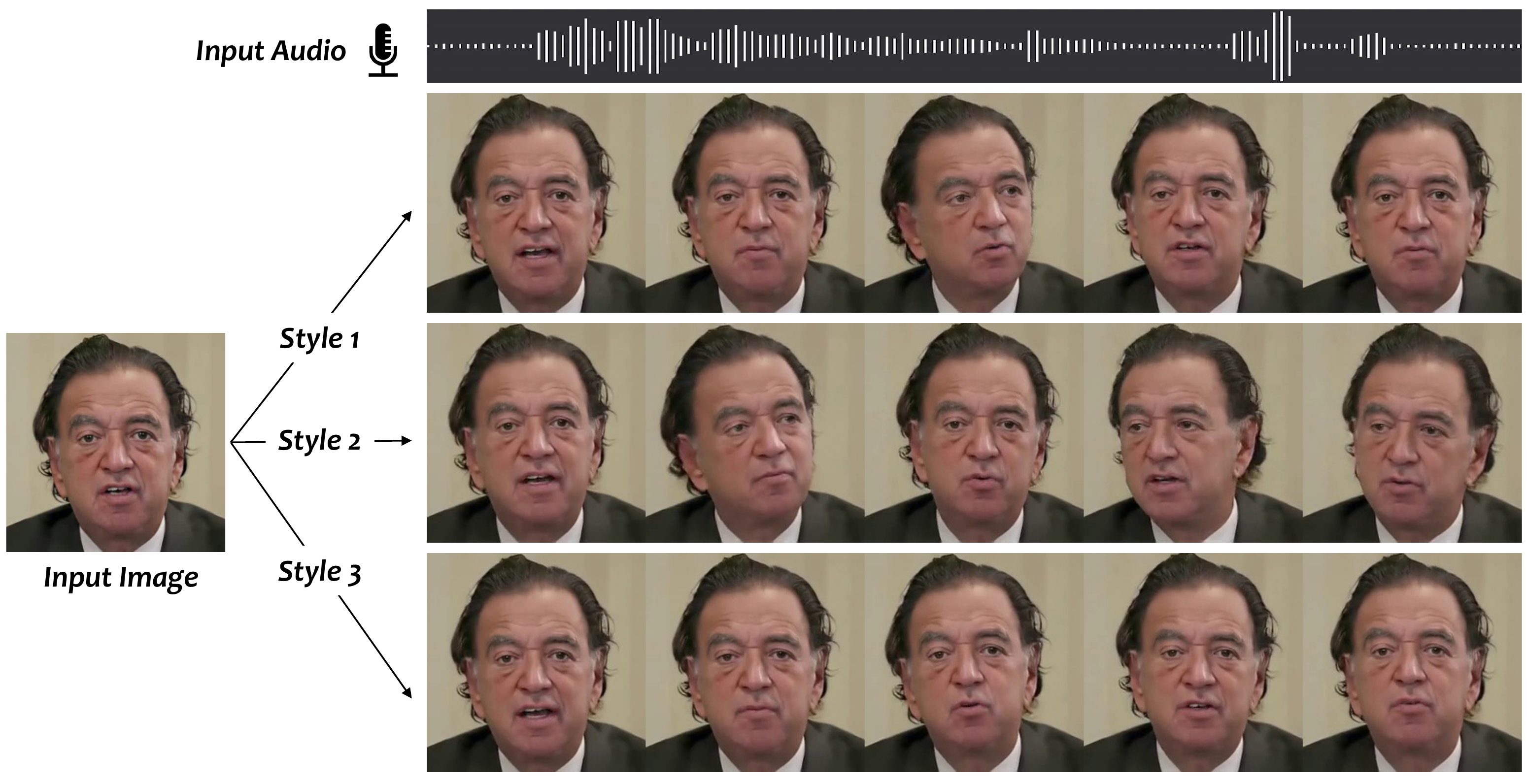}
    \vspace{-2em}
    \captionof{figure}{The proposed SadTalker produces diverse, realistic, synchronized talking videos from an input audio and a single reference image.}
\end{center}
}]

{
  \renewcommand{\thefootnote}%
    {\fnsymbol{footnote}}
  \footnotetext[1]{~Equal Contribution}
  \footnotetext[2]{~Corresponding Author}
}

\begin{abstract}
Generating talking head videos through a face image and a piece of speech audio still contains many challenges. \ie, unnatural head movement, distorted expression, and identity modification. We argue that these issues are mainly because of learning from the coupled 2D motion fields. On the other hand, explicitly using 3D information also suffers problems of stiff expression and incoherent video.
We present SadTalker, which generates 3D motion coefficients~(head pose, expression) of the 3DMM from audio and implicitly modulates a novel 3D-aware face render for talking head generation.
To learn the realistic motion coefficients, we explicitly model the connections between audio and different types of motion coefficients individually. Precisely, we present ExpNet to learn the accurate facial expression from audio by distilling both coefficients and 3D-rendered faces. 
As for the head pose, we design PoseVAE via a conditional VAE to synthesize head motion in different styles. 
Finally, the generated 3D motion coefficients are mapped to the unsupervised 3D keypoints space of the proposed face render, and synthesize the final video.
We conducted extensive experiments to demonstrate the superiority of our method in terms of motion and video quality.

\end{abstract}


\section{Introduction}
\label{sec:intro}
Animating a static portrait image with speech audio is a challenging task and has many important applications in the fields of digital human creation, video conferences, \etc. Previous works mainly focus on generating lip motion~\cite{atvgnet, lipgan, wav2lip, pcavs,videoretalking} since it has a strong connection with speech. Recent works also aim to generate a realistic talking face video containing other related motions, \eg, head pose. Their methods mainly introduce 2D motion fields by landmarks~\cite{zhou2020makelttalk} and latent warping~\cite{wang2021audio2head, wang2021one}. However, the quality of the generated videos is still unnatural and restricted by the preference pose~\cite{pcavs, eamm}, month blur~\cite{wav2lip}, identity modification~\cite{wang2021audio2head, wang2021one}, and distorted face~\cite{wang2021audio2head, wang2021one, hdtf}. 

Generating a natural-looking talking head video contains many challenges since the connections between audio and different motions are different. \ie, the lip movement has the strongest connection with audio, but audio can be talked via different head poses and eye blink. Thus, previous facial landmark-based methods~\cite{zhou2020makelttalk,atvgnet} and 2D flow-based audio to expression networks~\cite{wang2021one, wang2021audio2head} may generate the distorted face since the head motion and expression are not fully disentangled in their representation. Another popular type of method is the latent-based face animation~\cite{wav2lip, videoretalking, pcavs, eamm}. Their methods mainly focus on the specific 
kind of motions in talking face animation and struggle to synthesize high-quality video. Our observation is that the 3D facial model contains a highly decoupled representation and can be used to learn each type of motion individually. Although a similar observation has been discussed in \cite{hdtf}, their methods also generate inaccurate expressions and unnatural motion sequences.

From the above observation, we propose SadTalker, a \textbf{S}tylized \textbf{A}udio-\textbf{D}riven \textbf{Talk}ing-head video generation system through implicit 3D coefficient modulation. 
To achieve this goal, we consider the motion coefficients of the 3DMM as the intermediate representation and divide our task into two major components. On the one hand, we aim to generate the realistic motion coefficients~(\eg, head pose, lip motion, and eye blink) from audio and learn each motion individually to reduce the uncertainty. For expression, we design a novel audio to expression coefficient network by distilling the coefficients from the lip motion only coefficients from ~\cite{wav2lip} and the perceptual losses~(lip-reading loss~\cite{3dmm}, facial landmark loss) on the reconstructed rendered 3d face~\cite{deng2019accurate}.
For the stylized head pose, a conditional VAE~\cite{cvae} is used to model the diversity and life-like head motion by learning the residual of the given pose.
After generating the realistic 3DMM coefficients, we drive the source image through a novel 3D-aware face render. Inspired by face-vid2vid~\cite{facevid2vid}, we learn a mapping between the explicit 3DMM coefficients and the domain of the unsupervised 3D keypoint. Then, the warping fields are generated through the unsupervised 3D keypoints of source and driving and it warps the reference image to generate the final videos. We train each sub-network of expression generation, head poses generation and face renderer individually and our system can be inferred in an end-to-end style.
As for the experiments, several metrics show the advantage of our method in terms of video and motion methods.

The main contribution of this paper can be summarized as:

\begin{itemize}
    \item We present SadTalker, a novel system for a stylized audio-driven single image talking face animation using the generated realistic 3D motion coefficients.
    \item To learn the realistic 3D motion coefficient of the 3DMM model from audio, ExpNet and PoseVAE are presented individually.
    \item A novel semantic-disentangled and 3D-aware face render is proposed to produce a realistic talking head video.
    \item Experiments show that our method achieves state-of-the-art performance in terms of motion synchronization and video quality.
\end{itemize}

\section{Related Work}
\label{sec:related}

\paragraph{Audio-driven Single Image Talking Face Generation.} Early works~\cite{lipgan, wav2lip, videoretalking} mainly focus on producing accurate lip motion with a perception discriminator. Since the real videos contain many different motions, ATVGnet~\cite{atvgnet} uses the facial landmark as the intermediate representation to generate the video frames. A similar approach has been proposed by MakeItTalk~\cite{zhou2020makelttalk}, differently, it disentangles the content and speaker information from the input audio signal. 
Since facial landmarks are still a highly coupled space, generating the talking head in the disentangled space is also popular recently. PC-AVS~\cite{pcavs} disentangles the head pose and expression using implicit latent code. However, it can only produce low-resolution image and need the control signal from another video.
Audio2Head~\cite{wang2021audio2head} and Wang~\etal~\cite{wang2021one} get inspiration from the video-driven method~\cite{fomm} to produce the talking-head face. However, these head movements are still not vivid and produce distorted faces with inaccurate identities. Although there are some previous works~\cite{hdtf, ren2021pirenderer} use 3DMMs as an intermediate representation, their method still faces the problem of inaccurate expressions~\cite{ren2021pirenderer} and obvious artifacts~\cite{hdtf}.
 
\paragraph{Audio-driven Video Portrait.} Our task is also related to visual dubbing, which aims to edit a portrait video through audio. Different from audio-driven single image talking face generation, this task is typically required to be trained and edited on the specific video. Following previous work of deep video portrait~\cite{dvp}, these methods utilize 3DMM information for face reconstruction and animation. AudioDVP~\cite{audiodvp}, NVP~\cite{nvp}, AD-NeRF~\cite{adnerf} learn to reenact the expression to edit the mouth shape. Beyond lip movement, \ie, the head motions~\cite{facial, lsp}, emotional talking face~\cite{evp} also get attention. The 3DMM-based method plays an important role in these tasks since it is practical to fit the 3DMM parameters from a video clip. Although these methods achieve satisfactory results in personalized video, their method can not be applied to arbitrary photos and in the wild audio.

 \paragraph{Video-Driven Single Image Talking Face Generation.} This task is also known as face reenactment or face animation, which aims to transfer the motion of the source image to the target person. It has been widely explored~\cite{fomm, mraa, ren2021pirenderer, facevid2vid, fewshotvid2vid, styleheat, dagan, tps, lia, dpe} recently. Previous works also learn a shared intermediate motion representation from the source image and the target, which can be roughly divided into the landmark~\cite{fewshotvid2vid} and the unsupervised landmark-based methods~\cite{fomm, facevid2vid,dagan,tps}, 3DMM based methods~\cite{ren2021pirenderer, styleheat, doukas2020headgan} and the latent animation~\cite{lia, mallya2022implicit}. This task is much easier than our task since it contains the motion in the same domain. Our face render is also inspired by the method of unsupervised landmark-based method~\cite{facevid2vid} and 3DMM-based method~\cite{ren2021pirenderer} to map the learned coefficient to generate the real video. However, they are not focused on generating realistic motion coefficients.

\section{Method}
\label{sec:method}

\begin{figure}[tp]
    \centering
    \includegraphics[width=\columnwidth]{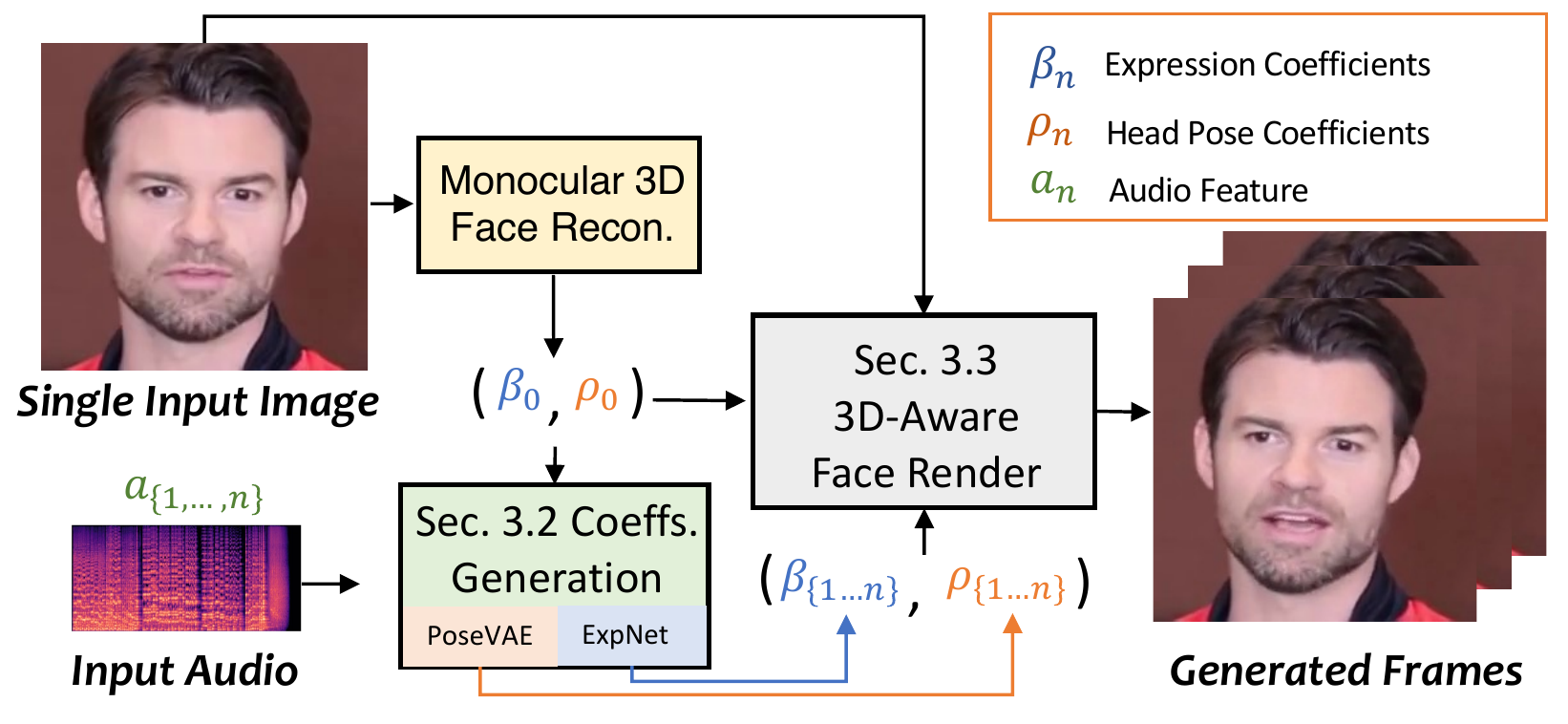}
    \vspace{-2em}
    \caption{Main pipeline. Our method uses the coefficients of 3DMM as intermediate motion representation. To this end, we first generate realistic 3D motion coefficients~(facial expression $\beta$, head pose $\rho$) from audio, then these coefficients are used to implicitly modulate the 3D-aware face render for final video generation.}
    \label{fig:main}
\end{figure}

As shown in Fig.~\ref{fig:main}, our system uses the 3D motion coefficients as the intermediate representation for talking head generation. We first extract the coefficients from the original image. Then, the realistic 3DMM motion coefficients are generated by ExpNet and PoseVAE individually. Finally, a 3D-aware face render is proposed to produce the talking head videos. Below, we give a brief introduction to the 3D face model as preliminaries in Sec.~\ref{sec:preliminary}, the audio-driven motion coefficients generation and the coefficients-driven image animator we design in Sec.~\ref{sec:audio2coeff} and Sec.~\ref{sec:render}, respectively.  

\subsection{Preliminary of 3D Face Model}
\label{sec:preliminary}
3D information is crucial to improve the realness of the generated video since the real video is captured in the 3D environment. However, previous works~\cite{pcavs, wav2lip, zhou2020makelttalk} have rarely been a consideration in 3D space since it is hard to obtain accurate 3D coefficients from a single image and the high-quality face render is also hard to design. 
Inspired by the recent single image deep 3D reconstruction method~\cite{deng2019accurate}, we consider the space of the predicted 3D Morphable Models~(3DMMs) as our intermediate representation.
In 3DMM, the 3D face shape $\mathbf{S}$ can be decoupled as:
\begin{equation}
\mathbf{S} = \overline{\mathbf{S}} + \mathbf{\alpha} \mathbf{U}_{id} + \mathbf{\beta} \mathbf{U}_{exp},
\end{equation}
where $\overline{\mathbf{S}}$ is the average shape of the 3D face, $\mathbf{U}_{id}$ and $\mathbf{U}_{exp}$ are the orthonormal basis of identity and expression of LSFM morphable model~\cite{3dmm}. 
Coefficients $\mathbf{\alpha} \in \mathbb{R}^{80}$ and $\mathbf{\beta} \in \mathbb{R}^{64}$ describe the person identity and expression, respectively. 
To preserve pose variance, coefficients $\mathbf{r} \in SO(3)$ and $\mathbf{t} \in \mathbb{R}^{3}$ denote the head rotation and translation. To achieve identity irrelevant coefficients generation~\cite{ren2021pirenderer}, we only model the parameters of motion as $ \left \{ \mathbf{\beta}, \mathbf{r}, \mathbf{t} \right \}$. We learn the head pose $ \rho = [\mathbf{r}, \mathbf{t}]$ and expression coefficients $\beta$ individually from the driving audio as introduced before. Then, these motion coefficients are used to implicitly modulate our face render for final video synthesis. 

\subsection{Motion Coefficients Generation through Audio}
\label{sec:audio2coeff}
As introduced above, the 3D motion coefficients contain both head pose and expression where the head pose is a global motion and the expression is relatively local. To this end, learning everything altogether will cause huge uncertainty in the network since the head pose has a relatively weak relationship with audio while the lip motion is highly connected.
We generate the motion of the head pose and expression using the proposed PoseVAE and ExpNet, respectively introduced below. 


\paragraph{ExpNet}

Learning a generic model which produces accurate expression coefficients from audio is extremely hard for two reasons: 1) audio-to-expression is not a one-to-one mapping task for different identities. 2) there are some audio-irrelevant motions in the expression coefficients and it will influence the prediction's accuracy. 
Our ExpNet is designed to reduce these uncertainties. As for the identity issue, we connect the expression motion to the specific person via the first frame's expression coefficients $\beta_0$. 
To reduce the motion weight of other facial components in natural talking, we use the \textit{lip motion only} coefficients as the coefficient target through the pre-trained network of Wav2Lip~\cite{wav2lip} and deep 3D reconstruction~\cite{deng2019accurate}. Then, other minor facial motions~(\eg, eye blink) can be leveraged via the additional landmark loss on the rendered images.

\begin{figure}[tp]
    \centering
    \includegraphics[width=\columnwidth]{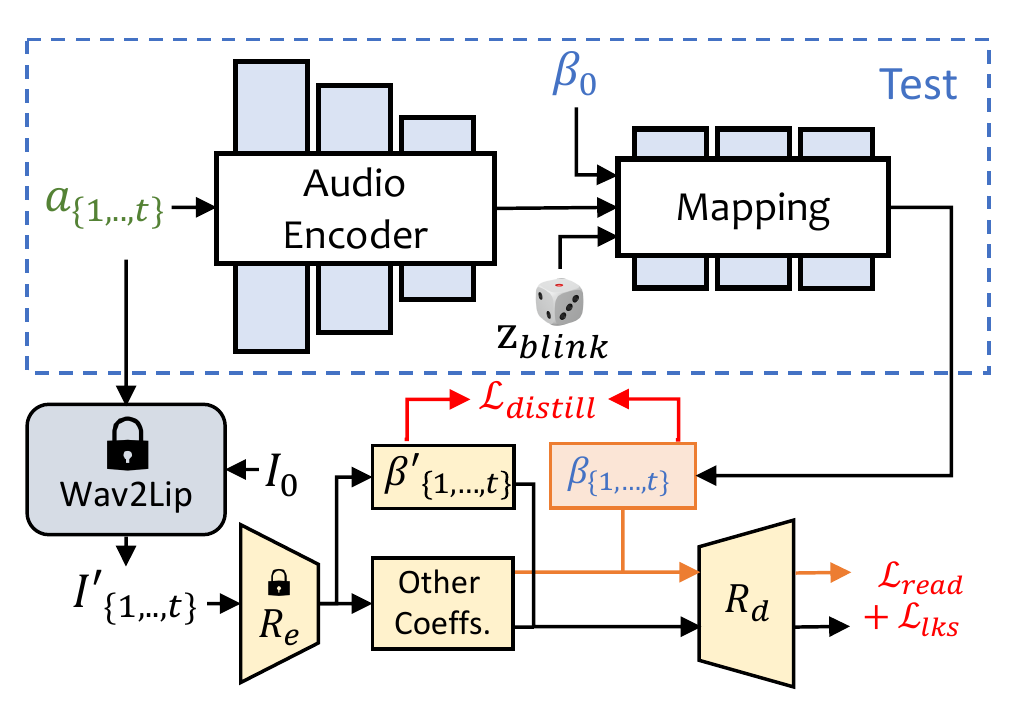}
    \vspace{-2.5em}
    \caption{The structure of our ExpNet. We involve a monocular 3D face reconstruction model~\cite{deng2019accurate}~($R_e$ and $R_d$) to learn the realistic expression coefficients. Where $R_e$ is a pretrained 3DMM coefficients estimator and $R_d$ is a differentiable 3D face render without learnable parameters. We use the reference expression $\beta_0$ to reduce the uncertainty of identity and the generated frame from pre-trained Wav2Lip~\cite{wav2lip} and the first frame as target expression coefficients since it only contains the lip-related motions. }
    \label{fig:expnet}
\end{figure}


As shown in Figure~\ref{fig:expnet}, we generate the $t$-frame expression coefficients from an audio window $a_{\{1,..,t\}}$, where the audio feature of each frame is a 0.2s mel-spectrogram. For training, we first design a ResNet-based audio encoder $\Phi_{A}$\cite{resnet, wav2lip} to embed the audio feature to a latent space. Then, a linear layer is added as the mapping network $\Phi_{M}$ to decode the expression coefficients. Here, we also add the reference expression $\beta_{0}$ from the reference image to reduce the identity uncertainty as discussed above. Since 
we use the lip-only coefficients as ground truth in the training, we explicitly add a blinking control signal $z_{blink} \in [0, 1]$ and the corresponding eye landmark loss to generate the controllable eye blinks. Formally, the network can be written as:
\begin{equation}
\beta_{\{1,...,t\}} = \Phi_{M}(\Phi_{A}(a_{\{1,...,t}\}), z_{blink}, \beta_{0})
\end{equation}

As for the loss function,
we first use $\mathcal{L}_{distill}$ to evaluate the differences between the lip only expression coefficients ${R}_{e}(\mathtt{Wav2Lip}(I_0, a_{\{1,...,t\}}))$ and the generated $\beta_{\{1,...,t\}}$. Notice that, we only use the first frame $I_{0}$ of the wav2lip to generate the lip-sync video which reduces the influence of the pose variant and  other facial expressions apart from lip movement. 
Besides, we also involve the differentiable 3D face render $R_{d}$ to calculate the additional perceptual losses in explicit facial motions space.  
As shown in Figure~\ref{fig:expnet}, we calculate the landmark loss $\mathcal{L}_{lks}$ to measure the range of eye blink and the overall expression accuracy. A pretrained lip reading network $\Phi_{reader}$ is also used as temporal lip reading loss $\mathcal{L}_{read}$ to keep the perceptual lip qualities~\cite{wav2lip, spectre}. We provide more training details in the supplementary materials.

\begin{figure}[tp]
    \centering
    \includegraphics[width=\columnwidth]{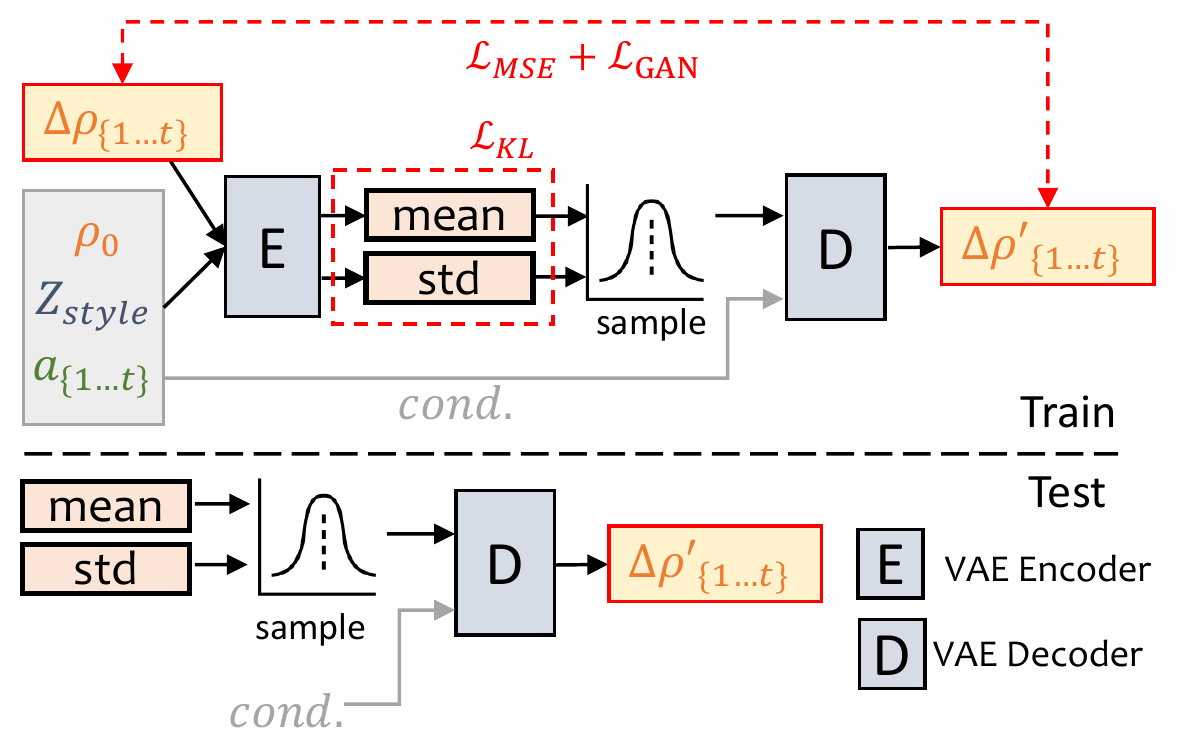}
    \vspace{-2em}
    \caption{The pipeline of the proposed PoseVAE. We learn the residual of the input head pose $\rho_{0}$ via a conditional VAE structure. Given the conditions: first frame $\rho_{0}$, style identity $Z_{style}$ and the audio clip $a_{\{1,...,t\}}$, our method learns a distribution of the residual head pose $\Delta\rho_{\{1,...,t\}} = \rho_{\{1,...,t\}} -\rho_{0}$. After training, we can generate the stylized results through the pose decoder and the conditions~$(cond.)$ only. 
    }
    \label{fig:posenet}
\end{figure}

\paragraph{PoseVAE}
As shown in Figure~\ref{fig:posenet}, a VAE~\cite{vae} based model is designed to learn the realistic and identity-aware stylized head movement $ \rho \in \mathbb{R}^6$ of the real talking video. In training, the pose VAE is trained on fixed $n$ frames using an encoder-decoder-based structure. Both the encoder and decoder are two-layer MLPs, where the inputs contain a sequential $t$-frame head poses and we embed it to a Gaussian distribution. In the decoder, the network is learned to generate the $t$-frame poses from the sampled distribution. Instead of generating the pose directly, our PoseVAE learns the \textit{residual} of the condition pose $\rho_0$ of the first frame, which enables our method to generate longer, stable, and continuous head motion in testing under the condition of the first frame. 
Besides, according to CVAE~\cite{cvae}, we add the corresponding audio feature $a_{\{1, ..., t\}}$ and style identity $Z_{style}$ as conditions for rhythm awareness and identity style. The KL-divergence $\mathcal{L}_{KL}$ is used to measure the distribution of the generated motions. The mean square loss $\mathcal{L}_{MSE}$ and adversarial loss $\mathcal{L}_{GAN}$ are used to ensure the generated quality. We provide more details about the loss function in the supplementary materials.


\subsection{3D-aware Face Render}
\label{sec:render}
\begin{figure}[tp]
    \centering
    \includegraphics[width=\columnwidth]{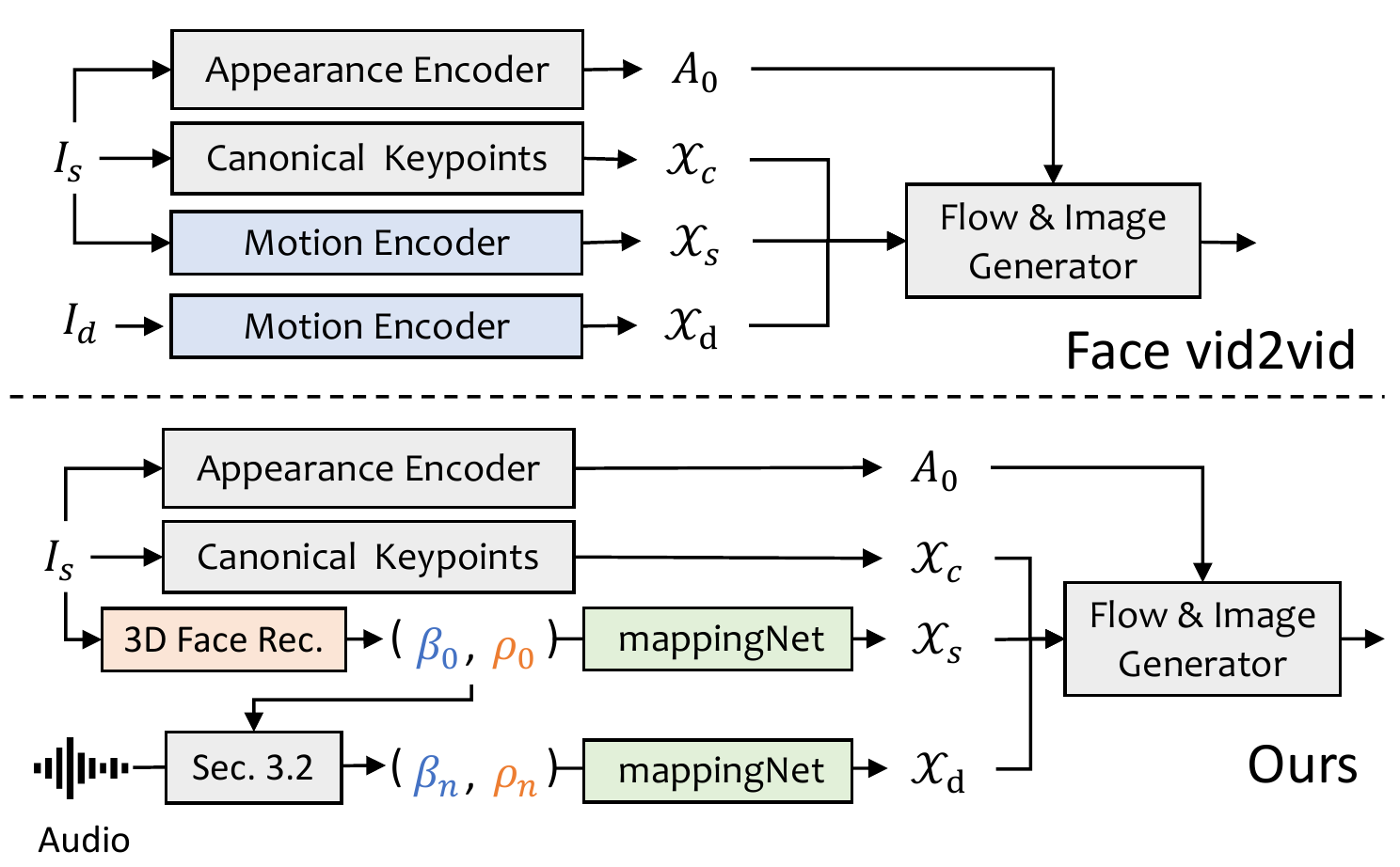}
    \vspace{-2em}
    \caption{The proposed FaceRender and comparison with face-vid2vid~\cite{facevid2vid}. Given source image $I_{s}$ and driving image $I_{d}$, face-vid2vid generates the motions in a unsupervised 3D keypoint spaces of $\mathcal{X}_{c}$, $\mathcal{X}_{s}$ and $\mathcal{X}_{d}$. Then, the image can be generated via the appearance $A_{0}$ and the keypoints. Since we do not have driving image, we use the explicit disentangled 3DMM coefficients as proxy and map it to the unsupervised 3D keypoints space.}
    \label{fig:render}
\end{figure}

After generating the realistic 3D motion coefficients, we render the final video through a well-designed 3D-aware image animator. We get inspiration from the recent image animation method face-vid2vid~\cite{facevid2vid} because it implicitly learns the 3D information from a single image. 
However, a real video is required as the motions driving signal in their method. Our face render makes it 
 drivable through 3DMM coefficients.
As shown in Figure~\ref{fig:render}, we propose mappingNet to learn the relationship between the explicit 3DMM motion coefficients~(head pose and expression) and the implicit unsupervised 3D keypoints. Our mappingNet is built via several 1D convolutional layers. We use the temporal coefficients from a time window for smoothing as PIRenderer\cite{ren2021pirenderer}. Differently, we find the face alignment motion coefficients in PIRenderer will hugely influence the motion naturalness of audio-driven video generation and provide an experiment in Sec.~\ref{ablation}. We only use the coefficients of expression and head pose.

As for training, our method contains two steps. Firstly, we train face-vid2vid~\cite{facevid2vid} in a self-supervised fashion as in the original paper. In the second step, we freeze all the parameters of the appearance encoder, canonical keypoints estimator, and image generator for tuning. Then, we train the mapping net on the 3DMM coefficients of the ground truth video in a reconstruction style. We give the supervision in the domain of unsupervised keypoints using $\mathcal{L}_1$ loss and the final generated video following their original implementation. More details can be founded in the supplementary materials.


\section{Experiments}
\label{Experiments}

\subsection{Implementation Details and Metrics}
\label{Details}



\paragraph{Datasets} We use VoxCeleb~\cite{voxceleb} dataset for training which contains over 100k videos of 1251 subjects. We crop the original videos following previous image animation methods~\cite{fomm} and resize the video to 256$\times$256. After preprocessing, the data is used to train our FaceRender. Since some videos and audios are not aligned in VoxCeleb, we select 1890 aligned videos and audios of 46 subjects to train our PoseVAE and ExpNet. The input audios are down-sampled to 16kHz and transformed to mel-spectrograms with the same setting as Wav2lip~\cite{wav2lip}. To test our method, we use the 346 videos' first 8-second video~(around 70k frames in total) from HDTF dataset~\cite{hdtf} since it contains high resolution and in-the-wild talking head videos. These videos are also cropped and processed following \cite{fomm} and resized to 256 $\times$256 for evaluation. We use the first frame of each video as the reference image to generate videos.

\paragraph{Implementation Details} All of ExpNet, PoseVAE, and FaceRender are trained separately and we employ Adam optimizer~\cite{adam} for all experiments. After training, our method can be inferred in an end-to-end fashion without any manual intervention. All the 3DMM parameters are extracted through pre-trained deep 3D face reconstruction method~\cite{deng2019accurate}. We perform all the experiments on 8 A100 GPUs. ExpNet, PoseVAE, and FaceRender are trained with a learning rate of $2e^{-5}$, $1e^{-4}$, and $2e^{-4}$, respectively. As for the temporal consideration, ExpNet uses continuous 5 frames to learn. PoseVAE is learned via  continuous 32 frames. The frames in FaceRender are generated frame-by-frame with the coefficients of 5 continuous frames for stability.

\paragraph{Evaluation Metrics} We demonstrate the superiority of our method on multiple metrics that have been widely used in previous studies. We employ Frechet Inception Distance (FID)~\cite{fid,fidpaper} and cumulative probability blur detection (CPBD)~\cite{cpbd} to evaluate the quality of the images, in which FID is for the realism of generated frames and CPBD is for the sharpness of generated frames. To evaluate identity preservation, we calculate the cosine similarity~(CSIM) of identity embedding between the source images and the generated frames, in which we use ArcFace~\cite{arcface} to extract identity embedding of images. 
To evaluate lip synchronization and mouth shape, 
we evaluate the perceptual differences of the mouth shape from Wav2Lip~\cite{wav2lip}, including the distance score~(LSE-D) and confidence score~(LSE-C). We also conduct some metrics to evaluate the head motions of generated frames. For the diversity of the generated head motions, a standard deviation of the head motion feature embeddings extracted from the generated frames using Hopenet~\cite{hopenet} is calculated. For the alignment of the audio and generated head motions, we compute Beat Align Score as in Bailando~\cite{bailando}. 

\begin{table*}[t]
\centering
\resizebox{\textwidth}{!}{%
\begin{tabular}{l|l|cc|cc|ccc}
\toprule
\multirow{2}{*}{Method} & \multirow{2}{*}{Eye Blink} & \multicolumn{2}{c|}{Lip Synchronization} & \multicolumn{2}{c|}{Learned Head Motion}  & \multicolumn{3}{c}{Video Quality} \\ \cline{3-9}
&  & LSE-C$\uparrow$ & LSE-D$\downarrow$ & Diversity$\uparrow$ & Beat Align$\uparrow$ & FID$\downarrow$ & CPBD$\uparrow$ & CSIM$\uparrow$ \\
\hline 
Real Video & N./A. & 8.211 & 6.982 & 0.259 & 0.271 & 0.000 & 0.428 &1.000 \\
\rowcolor{lightgray!30} Wav2Lip*~\cite{wav2lip} & N./A. & 10.221 &5.535  &N./A. & N./A. &21.725 &0.368 &0.849\\
\rowcolor{lightgray!30} PC-AVS**~\cite{pcavs} & from ref. &9.053 &6.355  & N./A. & N./A. &69.127 &0.206 &0.683\\
MakeItTalk~\cite{zhou2020makelttalk} & automatic &5.110 &10.059  &0.257 &0.268 &28.243 &0.283 &0.838\\
Audio2Head~\cite{wang2021audio2head} & automatic &7.357 &7.535   &0.181 &0.267 &24.392 &0.281 &0.823\\
Wang~\etal~\cite{wang2021one} & automatic &4.932 &10.055 &0.226 &0.268 &22.432 &0.295 &0.811 \\
Ours & controllable & 7.290 &7.772 & \textbf{0.278} & \textbf{0.293} & \textbf{22.057} & \textbf{0.335} & \textbf{0.843} \\
\bottomrule
\end{tabular}
}
\vspace{-1em}
\caption{
Comparison with the state-of-the-art method on HDTF dataset. We evaluate Wav2Lip~\cite{wav2lip} and PC-AVS~\cite{pcavs} in the one-shot settings. Wav2Lip* achieves the best video quality since it only animates the lip region while other regions are the same as the original frame. PC-AVS** is evaluated using the fixed reference pose and fails in some samples.
}
\vspace{-0.05in}
\label{tab:compare}
\end{table*}

\begin{figure*}[tp]
    \centering
    \includegraphics[width=\textwidth]{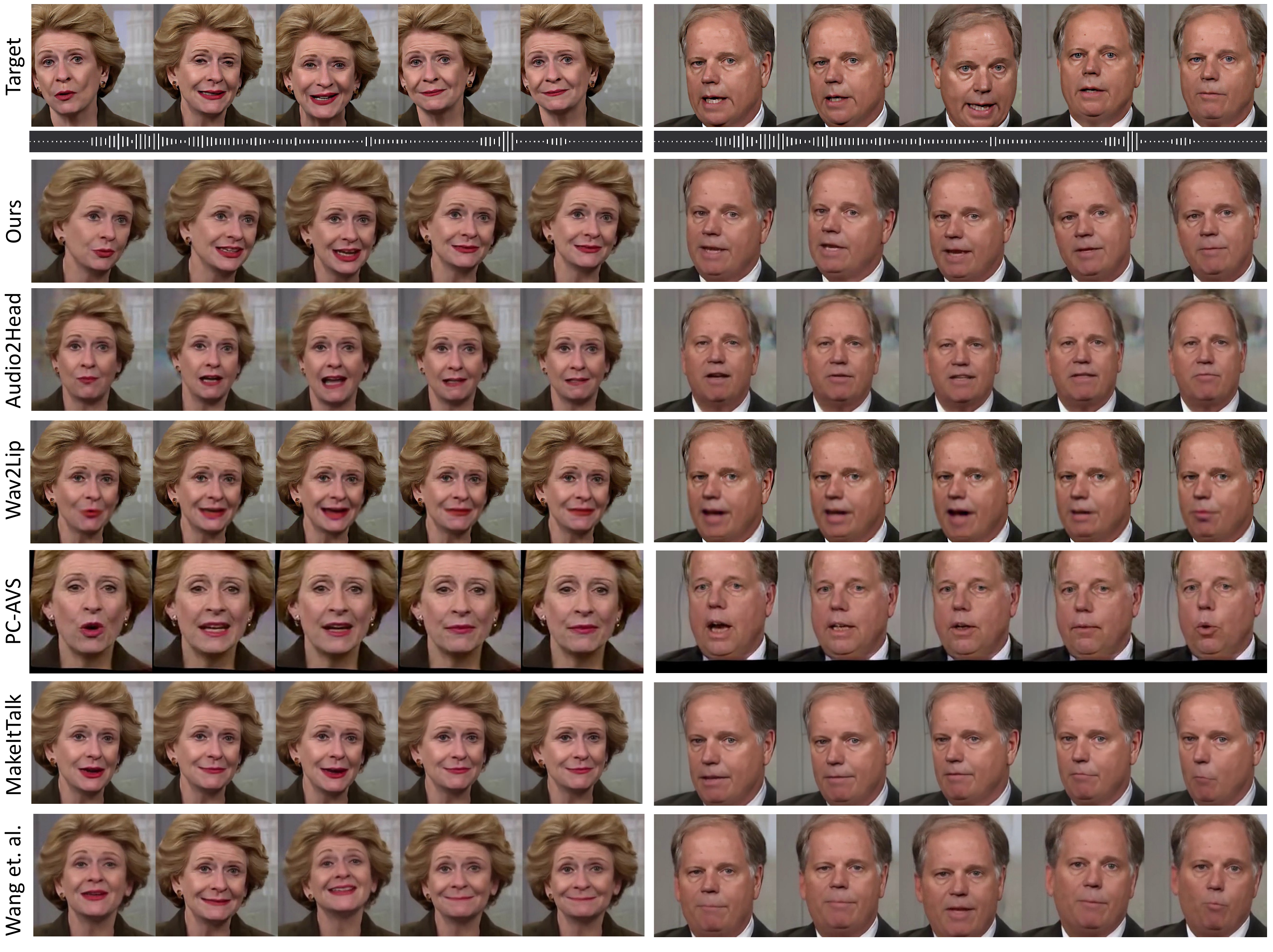}
    \vspace{-2em}
    \caption{We compare our method with several state-of-the-art methods for single image audio-driven talking head generation. Our method produces much higher quality results in terms of lip synchronization, identity preservation, head motion and image quality. \textit{We give the target image above for both lip shape and identity reference.} Please refer our supplementary video for better comparison. }
    \vspace{-1em}
    \label{fig:compare}
\end{figure*}

\subsection{Compare with other state-of-the-art methods}
We compare several state-of-the-art methods for the talking head video generations~(MakeItTalk~\cite{zhou2020makelttalk}, Audio2Head~\cite{wang2021audio2head} and Wang~\etal\cite{wang2021one}~\footnote{This method needs to extract the phoneme from audio, which only works on the specific language.}) and audio to expression generations~(Wav2Lip~\cite{wav2lip}, PC-AVS\cite{pcavs}). 
The evaluation is performed on their publicly available checkpoint directly. As shown in Table~\ref{tab:compare}, the proposed method shows better overall video qualities and head pose diversity and also shows comparable performance with other fully talking-head generation methods in terms of the lip synchronization metrics. We argue that these lip synchronization metrics are too sensitive to the audio where the unnatural lip movement may get a better score.
However, our method achieves a similar score to the real videos, which demonstrates our advantages. We also illustrate the visual results of different methods in Figure~\ref{fig:compare}. Here, we give the lip reference to visualize the lip synchronization of our method.
From the figure, our method has a very similar visual quality to the original target video and with different head poses as we expected. Compared with other methods, Wav2Lip~\cite{wav2lip} produces blur half-face. PC-AVS~\cite{pcavs} and Audio2Head~\cite{wang2021audio2head} are struggling for identity preservation. Audio2Head can only generate the front talking face. Besides, MakeItTalk~\cite{zhou2020makelttalk} and Audio2Head~\cite{wang2021audio2head} generate the distorted face video due to 2D warping. We give the video comparison in the supp. to show more clear comparisons.


\subsection{User Studies}
We conduct user studies to evaluate the performance of all the methods. We generate overall 20 videos as our testing. These samples contain almost equal genders with different ages, poses and expressions to show the robustness of our method. We invert 20 participants and let them choose the best method in terms of video sharpness, lip synchronization, the diversity and naturalness of the head motion, and overall quality. The results are shown in Table~\ref{tab:user}, where the participants like our method mostly because of the video and motion quality. We also find that 38\% of the participants think our methods show better lip synchronization than other methods, which is inconsistent with Table~\ref{tab:compare}. We think it might be because most of the participants focus on the overall quality of the video, where the blurry and still face videos~\cite{pcavs, wav2lip} influence their opinions.

\begin{table}[h]
\centering
\resizebox{\columnwidth}{!}{%
\begin{tabular}{l|cccc}
\toprule
Method & Lip & Motion & Video     & Overall  \\ 
& Sync. & Diversity   & Sharpness & Naturalness\\ \hline 
Wav2Lip~\cite{wav2lip} &15.6\% &3.1\%  &2.0\% &2.8\% \\
PC-AVS~\cite{pcavs} &18.1\% & 9.6\%  & 3.4\% & 9.1\%  \\
MakeItTalk~\cite{zhou2020makelttalk} & 5.6\% & 5.3\%  &5.7\% &6.9\%   \\
Wang~\etal~\cite{wang2021one} &12.5\% &12.1\%  &16.3\% &11.6\%  \\
Audio2Head~\cite{wang2021audio2head} &9.5\% &12.1\%  &9.7\% &14.7\%   \\
Ours & \textbf{38.7\%} & \textbf{57.9\%}  & \textbf{62.8\%} & \textbf{54.8\%} \\
\bottomrule
\end{tabular}}
\vspace{-1em}
\caption{User study.}
\vspace{-0.05in}
\label{tab:user}
\end{table}

\subsection{Ablation Studies}
\label{ablation}

\begin{figure}[tp]
    \centering
    \includegraphics[width=\linewidth]{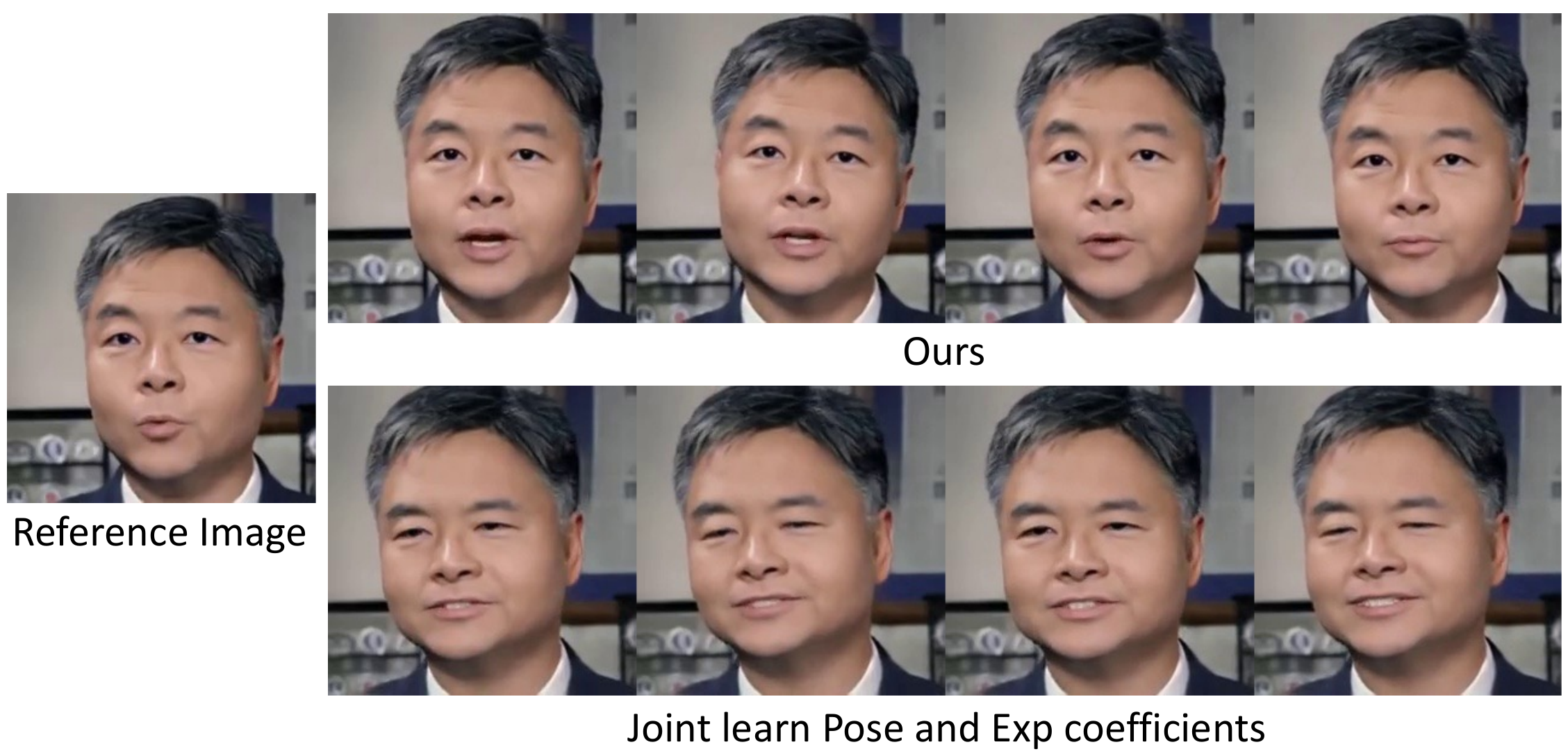}
    \vspace{-2em}
    \caption{We compare our method with a baseline method which learn all the coefficients from a single network without any condition~(from Speech2Gesture~\cite{speech2gesture}). Our method shows clear head movements, identity preservation and diverse expressions.}
    \label{fig:baseline}
\end{figure}




\begin{table}[t]
\centering
\resizebox{\columnwidth}{!}{%
\begin{tabular}{lccc}
\toprule
Method & LSE-C $\uparrow$ & LSE-D $\downarrow$  \\ \hline
Speech2Gesture~\cite{speech2gesture} &0.878 &13.889\\ \hline
OursFull~(Lip coeffs. + $\beta_0$ + $\mathcal{L}_{read}$) & \textbf{7.290} & \textbf{7.772}  \\
w/o $\beta_0$ \& $\mathcal{L}_{read}$ & 5.241 &9.532\\
w/o $\mathcal{L}_{read}$ & 6.993 &7.841 \\
w/ real coeffs. & 6.567 &8.061 \\
\bottomrule
\end{tabular}
}
\vspace{-1em}
\caption{Ablation for ExpNet. Both the initial expression $\beta_0$, lip reading loss $\mathcal{L}_{read}$ improve the performance a lot. However, the lip synchronization metric drops a lot when using the real coefficients.}
\vspace{-1em}

\label{tab:exp}
\end{table}

\begin{figure}[tp]
    \centering
    \includegraphics[width=\columnwidth]{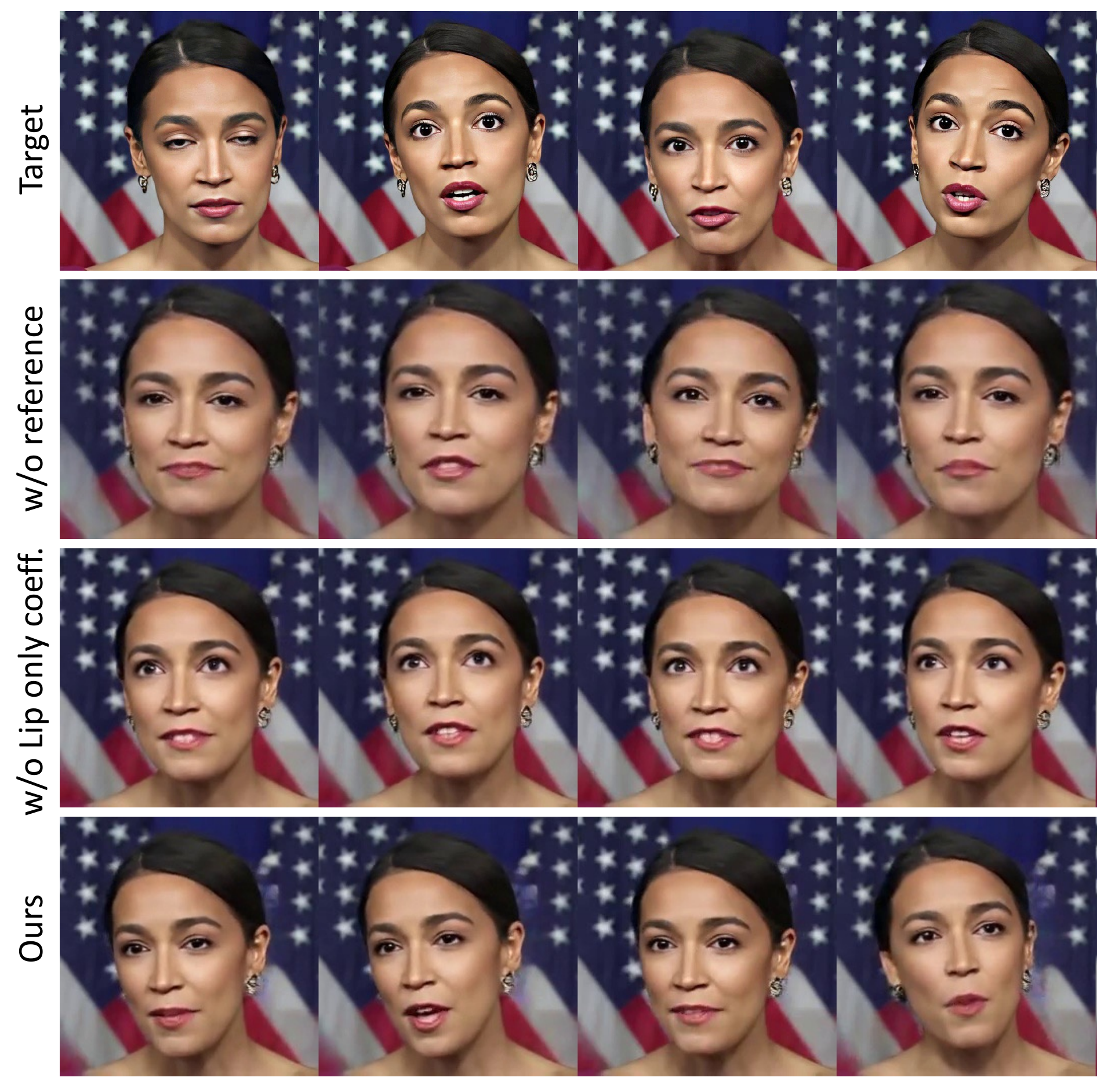}
    \vspace{-2em}
    \caption{The ablation of ExpNet. We choose four frames from the generated video as comparison. Our method largely reduces the uncertainty of audio to expression generation. The reference $\beta_0$ is used to provide the identity information while the lip only coefficients generate better lip synchronization. \textit{Notice that, the target image is provided as the identity and lip motion reference}.}
    \vspace{-1em}
    \label{fig:expnet_ablation}
\end{figure}

\paragraph{Ablation of ExpNet} For ExpNet, we mainly evaluate the necessity  of each component via the lip synchronization metrics. Since there are no disentangled methods before, we consider a baseline~(Speech2Gesture~\cite{speech2gesture}, which is an audio to keypoint generation network) to learn the head pose and expression coefficients jointly. As shown in Table~\ref{tab:exp} and Figure~\ref{fig:baseline}, learning all the motion coefficients altogether is hard to generate truth-worthy talking head videos. We then consider the variants of the proposed ExpNet, both the initial expression $\beta_0$, lip reading loss $\mathcal{L}_{read}$ and the necessity of lip-only coefficients are critical. The visual comparison is shown in Figure~\ref{fig:expnet_ablation}, where our method w/o the initial expression $\beta_0$ shows huge identity changes as expected. 
Also, if we use the real coefficients to replace the lip-only coefficients we use, the performance drops a lot in lip synchronization.

\paragraph{Ablation of PoseVAE}




We evaluate the proposed PoseVAE in terms of motion diversities and audio beat alignments. As shown in Table.~\ref{tab:pose}, the baseline Speech2Gesture~\cite{speech2gesture} also performs worse in pose evaluation. As for our variants, 
since our method contains several identity style labels, to better evaluate other components, we first consider the perform the ablation studies on a fixed one-hot style of our full method for evaluation~(OurFull, Single Fixed Style). 
Each condition in our settings benefits the overall motion quality in terms of diversity and beat alignment. We further report the results of the mixed style of our full method, which uses the randomly-selected identity label as style and shows a better diversity performance also. Since the pose differences are hard to be shown in the figure, please refer to our supplementary materials for better comparison.




\begin{table}[t]
\centering
\begin{tabular}{lcc}
\toprule
Method & Diversity$\uparrow$ & Beat Align$\uparrow$  \\ \hline
Speech2Gesture~\cite{speech2gesture}  & 0.1574 & 0.274  \\ \hline
OurFull~(Single Fixed Style) & 0.2735 & 0.287  \\
w/o  $\mathcal{L}_{gan}$  & 0.2500 & 0.271  \\
w/o  initial pose & 0.2725 & 0.278  \\
w/o  audio & 0.2566 &0.274  \\ 
w/o  all conditions & 0.2631 & 0.279  \\\hline
OursFull~(Mixed Style) & \textbf{0.2778} & \textbf{0.293} \\
\bottomrule
\end{tabular}
\vspace{-1em}
\caption{
Ablation the diversity and audio alignment of the proposed PoseVAE. Each component or conditional contribute largely to generate realistic head motions.
}
\vspace{-1em}
\label{tab:pose}
\end{table}

\begin{figure}[tp]
    \centering
    \includegraphics[width=\linewidth]{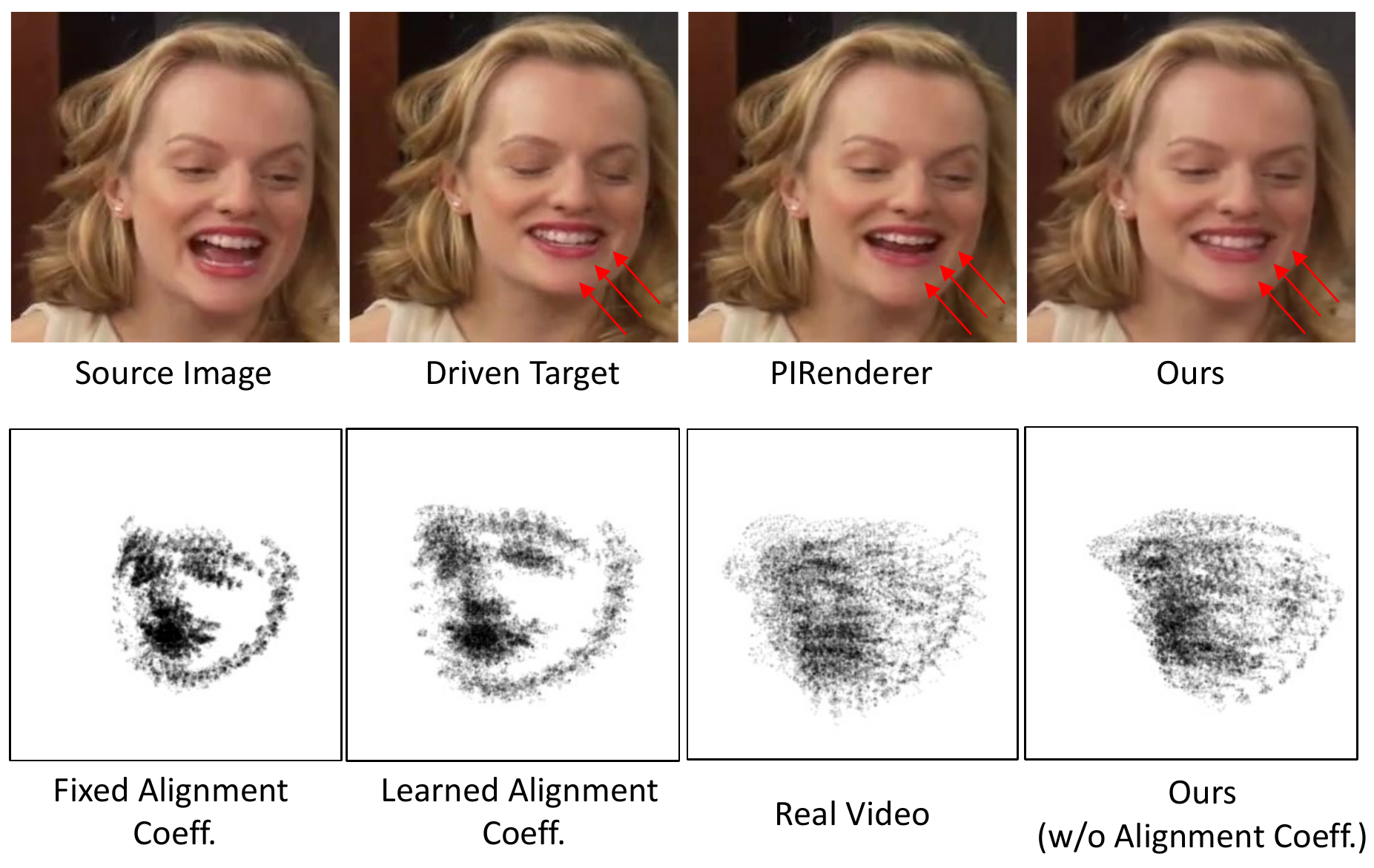}
    \vspace{-2em}
    \caption{Ablation studies of face render. In the first row, we directly compare our method with PIRenderer~\cite{ren2021pirenderer} for face animation and our method shows better expression modeling. The second row is the trace map of the generated facial landmarks from the same motion coefficients. Using additional face alignment coefficients as part of the motion coefficients~ \cite{ren2021pirenderer} will generate unrealistic aligned head video.}
    \label{fig:render_ablation}
\end{figure}

\paragraph{Ablation of Face Render}
We conduct the ablation study on the proposed face render from two aspects. On the one hand, we show the reconstruction quality of our method with the PIRenderer~\cite{ren2021pirenderer}, since both methods use 3DMM as an intermediate representation. As shown in the first row of Fig.~\ref{fig:render_ablation}, the proposed face render shows better expression reconstruction qualities thanks to the mapping of sparse unsupervised keypoints. Where accurate expression mapping is also the key to achieving lip synchronization. Besides, we evaluate the pose unnaturalness caused by the additional alignment coefficients used in PIRenderer~\cite{ren2021pirenderer}. As shown in the second row of Fig.~\ref{fig:render_ablation}, we plot the trace map of the landmarks from the generated video with the same head pose and expression coefficients. Using the fixed or learning-able crop coefficients~(as part of pose coefficients in our poseVAE) will generate the face-aligned video, which is strange as a natural video. We remove it and directly use the head pose and expression as modulation parameters showing a more realistic result.


\subsection{Limitation}
\begin{figure}[tp]
    \centering
    \includegraphics[width=\linewidth]{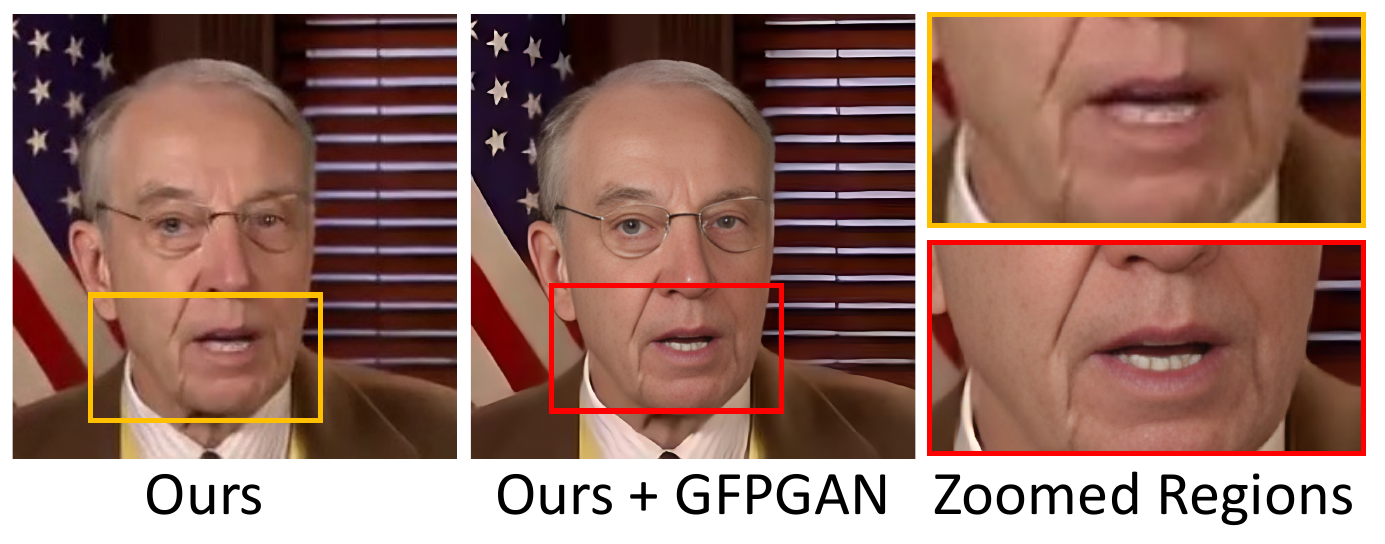}
    \vspace{-2em}
    \caption{Limitation. Our method may show some teeth artifacts in the lip region in some examples, it can be improved via the face restoration network, \ie, GFPGAN~\cite{gfpgan}.}
    \vspace{-1em}
    \label{fig:limitation}
\end{figure}

Although our method generates realistic video from a single image and audio, there still have some limitations in our system. Since 3DMMs do not model the variation of eyes and teeth, the mappingNet in our Face Render will also struggle to synthesize the realistic teeth in some cases. This limitation can be improved via the blind face restoration networks~\cite{gfpgan} as shown in Fig.~\ref{fig:limitation}. Another limitation of our work is that we only concern the lip motion and eye blinking other than the other facial expressions, \eg, emotion and gaze direction. Thus, the generated video has a fixed emotion, which also reduces the realism of generated content. We consider it as future work. 




\section{Conclusion}
\label{sec:conclusion}
In this paper, we present a new system for stylized audio-driven talking head video generation. We use the motion coefficients from 3DMM as an intermediate representation and learn the relationships. To generate realistic 3D coefficients from audio, we propose ExpNet and PoseVAE for realistic expressions and diverse head poses. To model the relationships between 3DMM motion coefficients and the real video, we propose a novel 3D-aware face render inspired by the image animation method~\cite{facevid2vid}. 
The experiments demonstrate the superiority of our entire framework. Since we predict the realistic 3D facial coefficients, our method can also be used in other modalities directly, \ie, personalized 2D visual dubbing~\cite{audiodvp}, 2D Cartoon animation~\cite{zhou2020makelttalk}, 3D face animation~\cite{faceformer,codetalker} and NeRF-based 4D talking-head generation~\cite{headnerf}.

\paragraph{Ethical Considerations} We consider the misuse of
the proposed method since it can generate very realistic video from a single face image. Both visible and invisible video watermarks will be inserted into the produced video for generated content identification similar to Dall-E~\cite{dalle} and Imagen~\cite{imagen}. We also hope our method can provide new research samples in the area of forgery detection.

{\small
\bibliographystyle{ieee_fullname}
\bibliography{11_references}

\begin{thebibliography}{10}\itemsep=-1pt

\bibitem{3dmm}
Volker Blanz and Thomas Vetter.
\newblock A morphable model for the synthesis of 3d faces.
\newblock In {\em ACM SIGGRAPH}, 1999.

\bibitem{atvgnet}
Lele Chen, Ross~K Maddox, Zhiyao Duan, and Chenliang Xu.
\newblock Hierarchical cross-modal talking face generation with dynamic
  pixel-wise loss.
\newblock In {\em CVPR}, 2019.

\bibitem{videoretalking}
Kun Cheng, Xiaodong Cun, Yong Zhang, Menghan Xia, Fei Yin, Mingrui Zhu, Xuan
  Wang, Jue Wang, and Nannan Wang.
\newblock Videoretalking: Audio-based lip synchronization for talking head
  video editing in the wild.
\newblock In {\em SIGGRAPH Asia 2022 Conference Papers}, pages 1--9, 2022.

\bibitem{arcface}
Jiankang Deng, Jia Guo, Niannan Xue, and Stefanos Zafeiriou.
\newblock Arcface: Additive angular margin loss for deep face recognition.
\newblock In {\em CVPR}, 2019.

\bibitem{deng2019accurate}
Yu Deng, Jiaolong Yang, Sicheng Xu, Dong Chen, Yunde Jia, and Xin Tong.
\newblock Accurate 3d face reconstruction with weakly-supervised learning: From
  single image to image set.
\newblock In {\em CVPR Workshops}, 2019.

\bibitem{cvae}
Carl Doersch.
\newblock Tutorial on variational autoencoders.
\newblock {\em arXiv preprint arXiv:1606.05908}, 2016.

\bibitem{doukas2020headgan}
Michail~Christos Doukas, Stefanos Zafeiriou, and Viktoriia Sharmanska.
\newblock Headgan: One-shot neural head synthesis and editing.
\newblock In {\em ICCV}, 2021.

\bibitem{faceformer}
Yingruo Fan, Zhaojiang Lin, Jun Saito, Wenping Wang, and Taku Komura.
\newblock Faceformer: Speech-driven 3d facial animation with transformers.
\newblock In {\em CVPR}, 2022.

\bibitem{spectre}
Panagiotis~P. Filntisis, George Retsinas, Foivos Paraperas-Papantoniou,
  Athanasios Katsamanis, Anastasios Roussos, and Petros Maragos.
\newblock Visual speech-aware perceptual 3d facial expression reconstruction
  from videos.
\newblock {\em arXiv preprint arXiv:2207.11094}, 2022.

\bibitem{speech2gesture}
Shiry Ginosar, Amir Bar, Gefen Kohavi, Caroline Chan, Andrew Owens, and
  Jitendra Malik.
\newblock Learning individual styles of conversational gesture.
\newblock In {\em CVPR}, 2019.

\bibitem{adnerf}
Yudong Guo, Keyu Chen, Sen Liang, Yong-Jin Liu, Hujun Bao, and Juyong Zhang.
\newblock Ad-nerf: Audio driven neural radiance fields for talking head
  synthesis.
\newblock In {\em ICCV}, 2021.

\bibitem{resnet}
Kaiming He, Xiangyu Zhang, Shaoqing Ren, and Jian Sun.
\newblock Deep residual learning for image recognition.
\newblock In {\em CVPR}, 2016.

\bibitem{fidpaper}
Martin Heusel, Hubert Ramsauer, Thomas Unterthiner, Bernhard Nessler, and Sepp
  Hochreiter.
\newblock Gans trained by a two time-scale update rule converge to a local nash
  equilibrium.
\newblock In {\em NeurIPS}, 2017.

\bibitem{dagan}
Fa-Ting Hong, Longhao Zhang, Li Shen, and Dan Xu.
\newblock Depth-aware generative adversarial network for talking head video
  generation.
\newblock In {\em CVPR}, 2022.

\bibitem{headnerf}
Yang Hong, Bo Peng, Haiyao Xiao, Ligang Liu, and Juyong Zhang.
\newblock Headnerf: A real-time nerf-based parametric head model.
\newblock In {\em CVPR}, 2022.

\bibitem{pix2pix}
Phillip Isola, Jun-Yan Zhu, Tinghui Zhou, and Alexei~A Efros.
\newblock Image-to-image translation with conditional adversarial networks.
\newblock {\em CVPR}, 2017.

\bibitem{eamm}
Xinya Ji, Hang Zhou, Kaisiyuan Wang, Qianyi Wu, Wayne Wu, Feng Xu, and Xun Cao.
\newblock Eamm: One-shot emotional talking face via audio-based emotion-aware
  motion model.
\newblock In {\em ACM SIGGRAPH}, 2022.

\bibitem{evp}
Xinya Ji, Hang Zhou, Kaisiyuan Wang, Wayne Wu, Chen~Change Loy, Xun Cao, and
  Feng Xu.
\newblock Audio-driven emotional video portraits.
\newblock In {\em CVPR}, 2021.

\bibitem{dvp}
Hyeongwoo Kim, Pablo Garrido, Ayush Tewari, Weipeng Xu, Justus Thies, Matthias
  Niessner, Patrick P{\'e}rez, Christian Richardt, Michael Zollh{\"o}fer, and
  Christian Theobalt.
\newblock Deep video portraits.
\newblock {\em ACM Transactions on Graphics (TOG)}, 2018.

\bibitem{adam}
Diederik~P Kingma and Jimmy Ba.
\newblock A method for stochastic optimization.
\newblock {\em arXiv preprint arXiv:1412.6980}, 2014.

\bibitem{vae}
Diederik~P Kingma and Max Welling.
\newblock Auto-encoding variational bayes.
\newblock {\em CoRR}, abs/1312.6114, 2014.

\bibitem{bailando}
Siyao Li, Yu Weijiang, Gu Tianpei, Lin Chunze, Wang Quan, Qian Chen, Loy~Chen
  Change, and Liu Ziwei.
\newblock Bailando: 3d dance generation via actor-critic gpt with choreographic
  memory.
\newblock In {\em CVPR}, 2022.

\bibitem{lsp}
Yuanxun Lu, Jinxiang Chai, and Xun Cao.
\newblock Live speech portraits: real-time photorealistic talking-head
  animation.
\newblock {\em ACM Transactions on Graphics (TOG)}, 2021.

\bibitem{ma2022training}
Pingchuan Ma, Yujiang Wang, Stavros Petridis, Jie Shen, and Maja Pantic.
\newblock Training strategies for improved lip-reading.
\newblock In {\em ICASSP}, 2022.

\bibitem{mallya2022implicit}
Arun Mallya, Ting-Chun Wang, and Ming-Yu Liu.
\newblock {Implicit Warping for Animation with Image Sets}.
\newblock In {\em NeurIPS}, 2022.

\bibitem{voxceleb}
Arsha Nagrani, Joon~Son Chung, and Andrew Zisserman.
\newblock Voxceleb: a large-scale speaker identification dataset.
\newblock In {\em INTERSPEECH}, 2017.

\bibitem{cpbd}
Niranjan~D. Narvekar and Lina~J. Karam.
\newblock A no-reference image blur metric based on the cumulative probability
  of blur detection (cpbd).
\newblock {\em TIP}, 2011.

\bibitem{hopenet}
Ruiz Nataniel, Eunji Chong, and Rehg~James M.
\newblock Fine-grained head pose estimation without keypoints.
\newblock In {\em CVPR Workshops}, 2018.

\bibitem{dpe}
Youxin Pang, Yong Zhang, Weize Quan, Yanbo Fan, Xiaodong Cun, Ying Shan, and
  Dong-ming Yan.
\newblock Dpe: Disentanglement of pose and expression for general video
  portrait editing.
\newblock {\em arXiv preprint arXiv:2301.06281}, 2023.

\bibitem{wav2lip}
K~R Prajwal, Rudrabha Mukhopadhyay, Vinay P.Namboodiri, and C.V.Jawahar.
\newblock A lip sync expert is all you need for speech to lip generation in the
  wild.
\newblock In {\em ACM MM}, 2020.

\bibitem{lipgan}
Prajwal~K R, Rudrabha Mukhopadhyay, Jerin Philip, Abhishek Jha, Vinay
  Namboodiri, and C~V Jawahar.
\newblock Towards automatic face-to-face translation.
\newblock In {\em ACM MM}, 2019.

\bibitem{dalle}
Aditya Ramesh, Prafulla Dhariwal, Alex Nichol, Casey Chu, and Mark Chen.
\newblock Hierarchical text-conditional image generation with clip latents.
\newblock {\em arXiv preprint arXiv:2204.06125}, 2022.

\bibitem{ren2021pirenderer}
Yurui Ren, Ge Li, Yuanqi Chen, Thomas~H Li, and Shan Liu.
\newblock Pirenderer: Controllable portrait image generation via semantic
  neural rendering.
\newblock In {\em ICCV}, 2021.

\bibitem{imagen}
Chitwan Saharia, William Chan, Saurabh Saxena, Lala Li, Jay Whang, Emily
  Denton, Seyed Kamyar~Seyed Ghasemipour, Burcu~Karagol Ayan, S~Sara Mahdavi,
  Rapha~Gontijo Lopes, et~al.
\newblock Photorealistic text-to-image diffusion models with deep language
  understanding.
\newblock {\em arXiv preprint arXiv:2205.11487}, 2022.

\bibitem{fid}
Maximilian Seitzer.
\newblock {pytorch-fid: FID Score for PyTorch}.
\newblock \url{https://github.com/mseitzer/pytorch-fid}, August 2020.
\newblock Version 0.2.1.

\bibitem{fomm}
Aliaksandr Siarohin, St{\'e}phane Lathuili{\`e}re, Sergey Tulyakov, Elisa
  Ricci, and Nicu Sebe.
\newblock First order motion model for image animation.
\newblock In {\em NeurIPS}, 2019.

\bibitem{mraa}
Aliaksandr Siarohin, Oliver Woodford, Jian Ren, Menglei Chai, and Sergey
  Tulyakov.
\newblock Motion representations for articulated animation.
\newblock In {\em CVPR}, 2021.

\bibitem{nvp}
Justus Thies, Mohamed Elgharib, Ayush Tewari, Christian Theobalt, and Matthias
  Nie{\ss}ner.
\newblock Neural voice puppetry: Audio-driven facial reenactment.
\newblock In {\em ECCV}, 2020.

\bibitem{wang2021audio2head}
Suzhen Wang, Lincheng Li, Yu Ding, Changjie Fan, and Xin Yu.
\newblock Audio2head: Audio-driven one-shot talking-head generation with
  natural head motion.
\newblock In {\em IJCAI}, 2021.

\bibitem{wang2021one}
Suzhen Wang, Lincheng Li, Yu Ding, and Xin Yu.
\newblock One-shot talking face generation from single-speaker audio-visual
  correlation learning.
\newblock In {\em AAAI}, 2022.

\bibitem{fewshotvid2vid}
Ting-Chun Wang, Ming-Yu Liu, Andrew Tao, Guilin Liu, Jan Kautz, and Bryan
  Catanzaro.
\newblock Few-shot video-to-video synthesis.
\newblock In {\em NeurIPS}, 2019.

\bibitem{facevid2vid}
Ting-Chun Wang, Arun Mallya, and Ming-Yu Liu.
\newblock One-shot free-view neural talking-head synthesis for video
  conferencing.
\newblock In {\em CVPR}, 2021.

\bibitem{gfpgan}
Xintao Wang, Yu Li, Honglun Zhang, and Ying Shan.
\newblock Towards real-world blind face restoration with generative facial
  prior.
\newblock In {\em CVPR}, 2021.

\bibitem{lia}
Yaohui Wang, Di Yang, Francois Bremond, and Antitza Dantcheva.
\newblock Latent image animator: Learning to animate images via latent space
  navigation.
\newblock {\em arXiv preprint arXiv:2203.09043}, 2022.

\bibitem{audiodvp}
Xin Wen, Miao Wang, Christian Richardt, Ze-Yin Chen, and Shi-Min Hu.
\newblock Photorealistic audio-driven video portraits.
\newblock {\em IEEE Transactions on Visualization and Computer Graphics},
  26(12):3457--3466, 2020.

\bibitem{codetalker}
Jinbo Xing, Menghan Xia, Yuechen Zhang, Xiaodong Cun, Jue Wang, and Tien-Tsin
  Wong.
\newblock Codetalker: Speech-driven 3d facial animation with discrete motion
  prior.
\newblock {\em arXiv preprint arXiv:2301.02379}, 2023.

\bibitem{styleheat}
Fei Yin, Yong Zhang, Xiaodong Cun, Mingdeng Cao, Yanbo Fan, Xuan Wang, Qingyan
  Bai, Baoyuan Wu, Jue Wang, and Yujiu Yang.
\newblock Styleheat: One-shot high-resolution editable talking face generation
  via pre-trained stylegan.
\newblock In {\em ECCV}, 2022.

\bibitem{facial}
Chenxu Zhang, Yifan Zhao, Yifei Huang, Ming Zeng, Saifeng Ni, Madhukar
  Budagavi, and Xiaohu Guo.
\newblock Facial: Synthesizing dynamic talking face with implicit attribute
  learning.
\newblock In {\em ICCV}, 2021.

\bibitem{hdtf}
Zhimeng Zhang, Lincheng Li, Yu Ding, and Changjie Fan.
\newblock Flow-guided one-shot talking face generation with a high-resolution
  audio-visual dataset.
\newblock In {\em CVPR}, 2021.

\bibitem{tps}
Jian Zhao and Hui Zhang.
\newblock Thin-plate spline motion model for image animation.
\newblock In {\em CVPR}, 2022.

\bibitem{pcavs}
Hang Zhou, Yasheng Sun, Wayne Wu, Chen~Change Loy, Xiaogang Wang, and Ziwei
  Liu.
\newblock Pose-controllable talking face generation by implicitly modularized
  audio-visual representation.
\newblock In {\em CVPR}, 2021.

\bibitem{zhou2020makelttalk}
Yang Zhou, Xintong Han, Eli Shechtman, Jose Echevarria, Evangelos Kalogerakis,
  and Dingzeyu Li.
\newblock Makelttalk: speaker-aware talking-head animation.
\newblock {\em ACM Transactions on Graphics (TOG)}, 2020.

\end{thebibliography}
}

\ifarxiv \clearpage \appendix
\label{sec:appendix}
\renewcommand\thetable{\Alph{section}\arabic{table}}   
\renewcommand\thefigure{\Alph{section}\arabic{figure}}  






\section{Additional Experiments}
\label{sec:A}

\subsection{PIRenderer \emph{v.s} Our FaceRender for Face Reenactment}
We compare our FaceRender and PIRenderer~\cite{ren2021pirenderer} on the task of video-driven face reenactment. We have already shown the visual comparisons in Fig~\ref{fig:render_ablation} of the main paper, here, we give the numerical comparison results on the HDTF dataset. We evaluate these two methods using cross-identity settings and the results are conducted over 354 videos. As shown in Tab~\ref{tab:pirender}, the proposed method shows a much better visual quality in terms of FID and CSIM, which demonstrate the advantage of the proposed methods for audio-driven talking-head generation. For more differences between the proposed method and PIRenderer for this task, we also discuss the influence of the face alignment coefficient in Sec.~\ref{supp:align}.

\begin{table}[h]
\centering
\begin{tabular}{lccc}
\toprule
Method & FID $\downarrow$ & CPBD $\uparrow$ & CSIM $\uparrow$  \\ \hline
PIRenderer~\cite{ren2021pirenderer} &26.521 &\textbf{0.363} & 0.857\\ \hline
Our face render & \textbf{19.646} & 0.334 & \textbf{0.880}  \\ 
\bottomrule
\end{tabular}
\caption{Face render evaluation.}
\label{tab:pirender}
\end{table}



\subsection{Cross-ID Settings and More Test Datasets}
\label{sec:datasets}
We conduct the \textit{same-identity} experiment in the main paper, where the first frame of the test video is regarded as the reference image and the corresponding audio is regarded as the driving signal, generating a video that has the synchronized expression but diverse head poses. Differently, in the cross-identity experiment, the driving audio comes from another video. This kind of setting is also widely used in the comparison of the video-driven face reenactment~\cite{fomm}. In the cross-identity experiment, the reference pose of PC-AVS comes from the audio's corresponding video.

To this end, besides the HDTF~\cite{hdtf} dataset in the main paper~(we evaluate the results under the same-identity experiment), we also evaluate our method on HDTF and VoxCeleb2~\cite{voxceleb} datasets in \textbf{\textit{cross-identity setting}} as in Table~\ref{tab:hdtf_crossid} and Table~\ref{tab:vox_crossid}. VoxCeleb2 contains over 1 million utterances of 6112 speakers, of which there are 36k utterances of 118 speakers in the test set. We randomly select 3 videos for each speaker, obtaining 354 videos as for evaluation. The evaluation metrics are the same as those the same-identity experiment on the HDTF dataset. We directly evaluate the pretrained model of all the models on this dataset. We crop the videos in the same way used in~\cite{fomm} and resize the frames to 256$\times$256. As shown in the Tables, the proposed method shows 
 better lip synchronization in this kind of setting on both datasets in most metrics. The same trend is also observed in the head motion and visual quality of the final videos.

\begin{table*}[t]
\centering
\resizebox{\textwidth}{!}{%
\begin{tabular}{l|cc|cc|ccc}
\toprule
\multirow{2}{*}{Method} & \multicolumn{2}{c|}{Lip Synchronization} & \multicolumn{2}{c|}{Learned Head Motion}  & \multicolumn{3}{c}{Video Quality} \\ \cline{2-8}
 & LSE-C$\uparrow$ & LSE-D$\downarrow$ & Diversity$\uparrow$ & Beat Align$\uparrow$ & FID$\downarrow$ & CPBD$\uparrow$ & CSIM$\uparrow$  \\
\hline 
Real Video  & 8.211 & 6.982 & 0.259 & 0.271 & 0 & 0.428 & 1.000 \\
\rowcolor{lightgray!30} Wav2Lip*~\cite{wav2lip}  & 9.641 &6.035  &N./A. & N./A. &21.727 &0.368 &0.846 \\
\rowcolor{lightgray!30} PC-AVS**~\cite{pcavs}  &8.959 &6.435  & N./A. & N./A. &99.098 &0.201 &0.648 \\
MakeItTalk~\cite{zhou2020makelttalk}  &4.937 &10.231  &0.2553 &0.276 &26.829 &0.333 &0.834 \\
Audio2Head~\cite{wang2021audio2head} &7.237 & \bf7.648   &0.1783 &0.260 &24.404 &0.282 &0.818 \\
Wang~\etal~\cite{wang2021one} &4.634 &10.457 &0.2260 &0.265 &22.302 &0.294 &0.805  \\
Ours & \textbf{7.343} & 7.709 & \textbf{0.2759} & \textbf{0.284} &\textbf{20.886} & \textbf{0.334} & \textbf{0.846}  \\
\bottomrule
\end{tabular}
}
\caption{
Comparison with the state-of-the-art method on HDTF dataset~\cite{hdtf} with \textit{cross-identity} setting.   Wav2Lip* achieves the best video quality since it only animates the lip region while other regions are the same as the original frame. In cross-identity setting, PC-AVS** is evaluated using the reference pose from the driving video and fails in some samples.
}

\label{tab:hdtf_crossid}
\end{table*}

\begin{table*}[t]
\centering
\resizebox{\textwidth}{!}{%
\begin{tabular}{l|cc|cc|ccc}
\toprule
\multirow{2}{*}{Method} & \multicolumn{2}{c|}{Lip Synchronization} & \multicolumn{2}{c|}{Learned Head Motion}  & \multicolumn{3}{c}{Video Quality} \\ \cline{2-8}
 & LSE-C$\uparrow$ & LSE-D$\downarrow$ & Diversity$\uparrow$ & Beat Align$\uparrow$ & FID$\downarrow$  & CPBD$\uparrow$ & CSIM$\uparrow$  \\
\hline 
Real Video  & 6.209 & 7.911 & 0.4879 & 0.266 &0 & 0.099 & 1.000 \\
\rowcolor{lightgray!30} Wav2Lip*~\cite{wav2lip}  & 7.640 &7.099  &N./A. & N./A. &19.293 &0.107 &0.936 \\
\rowcolor{lightgray!30} PC-AVS**~\cite{pcavs}  &7.168 &7.443  & N./A. & N./A. &111.043 &0.074 &0.494 \\
MakeItTalk~\cite{zhou2020makelttalk}  &3.756 &10.222  &\textbf{0.5230} &0.275 &23.501 &0.063 &0.883 \\
Audio2Head~\cite{wang2021audio2head} &5.266 &8.788   &0.2064 &0.273 &54.694 &0.098 &0.602 \\
Wang~\etal~\cite{wang2021one} &3.441 &10.519 &0.2547 &0.272 &42.092 &\textbf{0.136} &0.750  \\
Ours & \textbf{5.571} &\textbf{8.503} & 0.5211 & \textbf{0.277} &\textbf{22.738} & 0.081 & \textbf{0.893}  \\
\bottomrule
\end{tabular}
}

\caption{
Comparison with the state-of-the-art method on VoxCeleb2~\cite{voxceleb} dataset under \textit{cross-identity} setting.   Wav2Lip* achieves the best video quality since it only animates the lip region while other regions are the same as the original frame. In cross-identity setting, PC-AVS** is evaluated using the reference pose from the driving video and fails in some samples.
}

\label{tab:vox_crossid}
\end{table*}

\section{More Implementation Details}
We provide the detailed audio pre-processing, network structures, loss function, the discussion on alignment coefficients in Sec.~\ref{supp:audio}, Sec.~\ref{supp:network}, Sec.~\ref{supp:loss} and Sec.~\ref{supp:align}.

\subsection{Audio Pre-processing Details}
\label{supp:audio}
We follow Wav2Lip~\cite{wav2lip} to pre-process the audio. Specifically, we pre-process all the audio to 16k Hz. Then, we convert it to
the mel-spectrograms with FFT window size 800, hop length
200 and 80 Mel filter banks. Thus, for each frame, we have 0.2s mel-spectrogram feature with the shape of 16$\times$80.

\subsection{Network Structure Details}
\label{supp:network}

\begin{figure*}[tp]
    \centering
    \includegraphics[width=\textwidth]{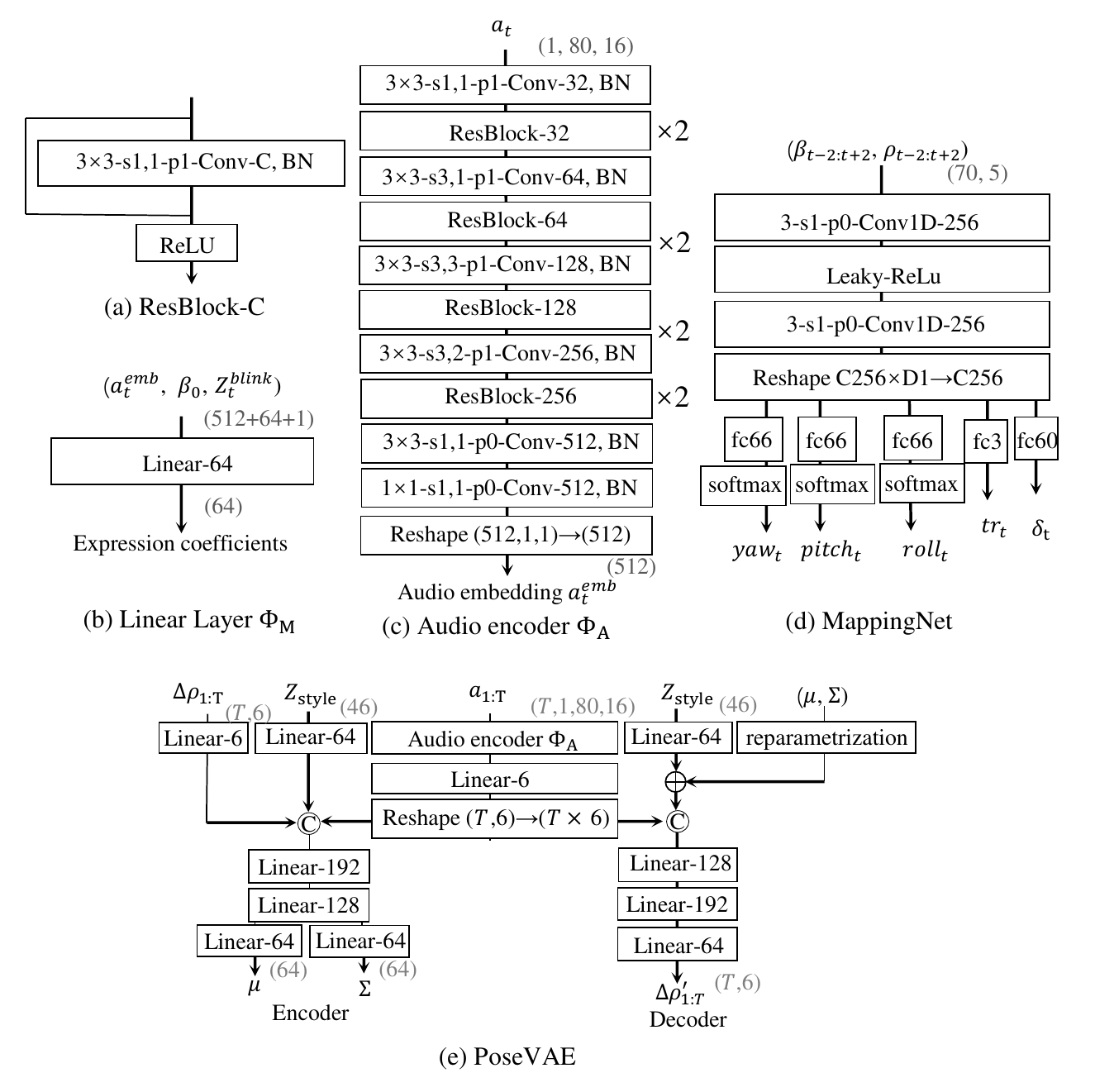}
    \vspace{-2em}
    \caption{The architectures of the networks in our model. Here,  `3×3-s1,1-p1-Conv-32' means a convolutional layer with the kernel size 3$\times$3, the stride size (1,1), padding size (1,1) and the output channel is 32.}
    \vspace{-1em}
    \label{fig:architecture}
\end{figure*}


\paragraph{ExpNet} Our ExpNet is built via an audio encoder $\Phi_{A}$ and a linear layer $\Phi_{M}$. We use the parameters from the pre-trained Wav2Lip to initialize the audio encoder. As discussed in the main paper, we first encode the audio feature into an audio embedding. And then, we generate the expression coefficients $\beta^g_{\{1,...,t\}}$ with additional coefficients of the first frame $\beta_0$ and blink control signal~$z_{blink}$. As shown in Fig.~\ref{fig:architecture}~(c), the audio encoder $\Phi_{A}$ is built via a four stages ResBlock-C as in Fig.~\ref{fig:architecture}~(a). we only use a single linear layer $\Phi_{M}$ as in Fig.~\ref{fig:architecture}~(b).

\paragraph{PoseVAE} As shown in Fig.~\ref{fig:architecture}~(e), both the encoder and decoder of our PoseVAE contain several linear layers. For the conditions, the encoder $\mu$ and $\sum$ is mapped through the concatenation of the $\Delta \rho_{\{1:T\}}$, the 46 dimensional one-hot vector $Z_{style}$~(our training dataset contains 46 identities) and the encoded features from audio encoder $\Phi_{A}$. As for the decoder, we first add the re-parameterized feature and the style embedding. Then, we concatenate the audio feature similar to the encoder.

\paragraph{FaceRender} As discussed in the main paper of Fig.~\ref{fig:render}, our face render is inspired by the motion transfer method face-vid2vid~\cite{facevid2vid}. We introduce a mappingNet to remap the learned 3DMM motion coefficients to the space of unsupervised 3D keypoints. As shown in Figure~\ref{fig:architecture}~(d), the mapping network contains the $t$-frames~($[t-2:t+2]$) motion coefficients of pose $\rho_{[t-2:t+2]}$ and expression $\beta_{[t-2:t+2]}$ to generate the motions representation of face-vid2vid~\cite{facevid2vid}~( $\texttt{yaw}$, $\texttt{pitch}$, $\texttt{roll}$, $\texttt{tr}$, and $\delta$) in frame $t$. Other networks in our FaceRender have the same structures in \cite{facevid2vid}. Please refer to face-vid2vid~\cite{facevid2vid} for more network details on the FaceRender.

\subsection{Loss Function Details}
\label{supp:loss}

\paragraph{ExpNet} As described in the main paper, we use the expression coefficients which are generated from the pre-trained wav2lip~\cite{wav2lip} and then perform 3D face capture~\cite{deng2019accurate} as guidance~(lip-only expression coefficients for short). Basically, we calculate $\mathcal{L}_{distill}$ through the Mean-Squared loss between lip-only expression coefficients and the generated expression coefficients in training. Formally, a $T$-frames $\mathcal{L}_{distill}$ can be written as:
\begin{equation}
\mathcal{L}_{distill}=\frac{1}{T} \sum_{t=1}^{T}\left({\beta}_{t}^g - {\beta}_{t}^{lip}\right)^{2}
\end{equation}
Where ${\beta}_{t}^{lip} $ and  ${\beta}_{t}^g$ are the lip only and the generated expression coefficients, respectively.

We also calculate the loss function on the projected 2D landmarks of the rendered 3D face. In detail, as shown in Fig.~\ref{fig:lms}, the height and width of the eye area in the $t$-th frame are defined as follows:
\begin{align}
{E}^{w}_{t}=\frac{\left\|P^{39}_t-P^{36}_t \right\|_{2} + \left\|P^{45}_t-P^{42}_t \right\|_{2} }{2} \\
{E}^{h}_{t}=\frac{\left\|P^{37}_t+P^{38}_t-P^{40}_t-P^{41}_t \right\|_{2}}{2} \\ + 
\frac{\left\|P^{43}_t+P^{44}_t-P^{46}_t-P^{47}_t \right\|_{2} }{2}.
\end{align}
Where ${E^{w}_{t}}$ is the width of the eye area in frame ${t}$, ${E}^{h}_{t}$ is the width of the eye area in frame ${t}$, $P^{i}_t$ is the ${i}$-th landmark in frame ${t}$. we define ${R_t} = \frac{{E}^{h}_{t}}{{E}^{w}_{t}}$ as the predicted and calculate eye loss as follows:
\begin{equation}
\mathcal{L}_{eye}=\sum_{t=1}^{T} \left\|{R_t}-Z^{blink}_{t} \right\|_{1},
\end{equation}
where $Z^{blink}_{t}$ is the eye blinking control signal the of ${t}$-th frame which is generated uniformly and randomly. To eliminate the effects of $\mathcal{L}_{eye}$ on other facial expression, we also constrain the minimal modification in the other landmarks. Thus, 
\begin{equation}
    \mathcal{L}_{lks}= \lambda_{eye} \mathcal{L}_{eye} + \frac{1}{T} \frac{1}{N}  \sum_{t=1}^{T} \sum_{i=1}^{M} \left\|P^{i}_t-P^{i\prime}_{t}\right\|^{2}_{2},
\end{equation}
where $\lambda_{eye}$ is set to 200, $P^{i\prime}_{t}$ is the landmarks predicted by the lip-only expression coefficients, $\mathcal{M}$ is a set of landmarks other than the eye areas.

\begin{figure}[tp]
    \centering
    \includegraphics[width=0.5\columnwidth]{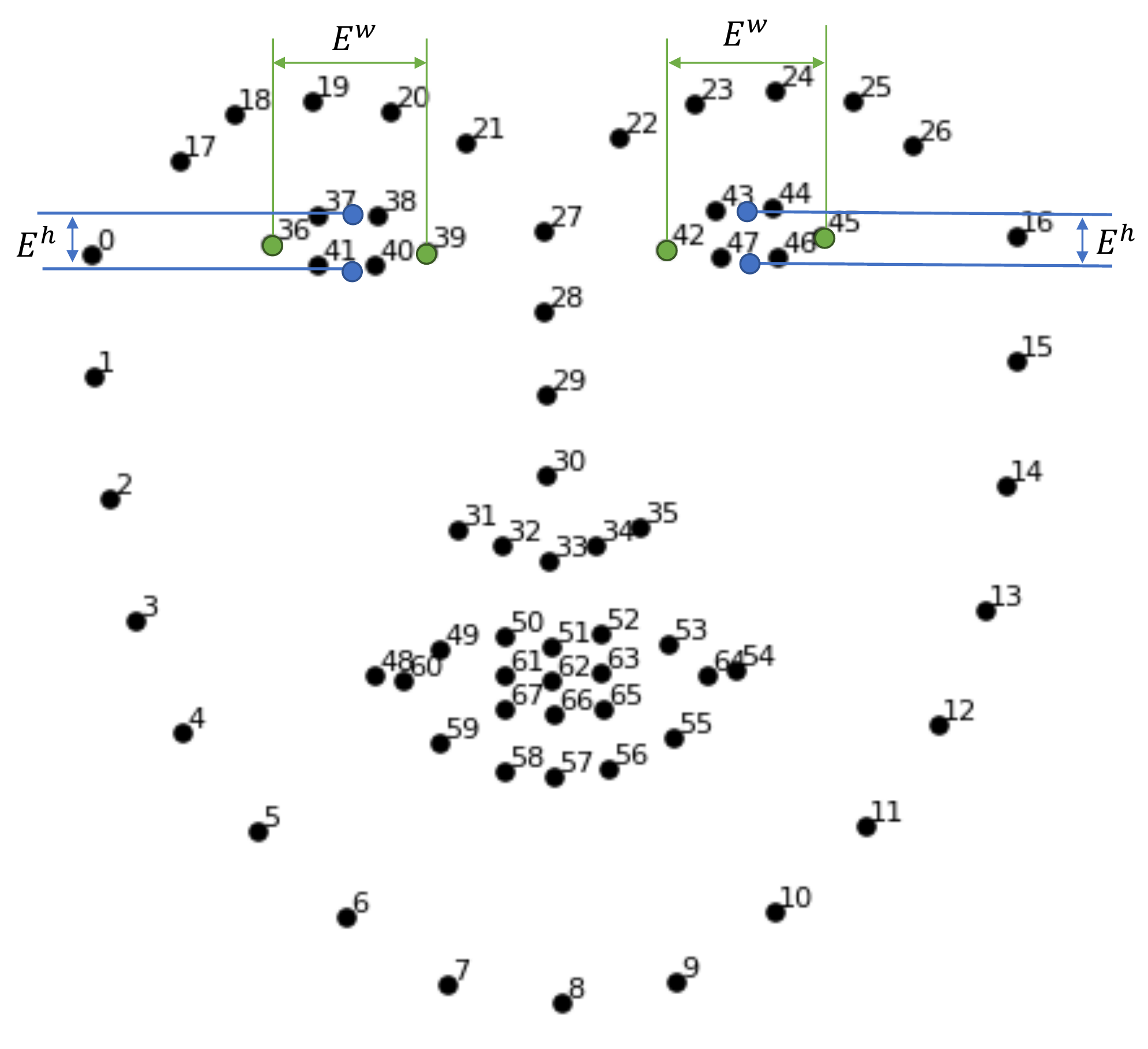}
    \caption{Face landmarks visualization.}
    \label{fig:lms}
\end{figure}

\begin{figure}[tp]
    \centering
    \includegraphics[width=0.6\linewidth]{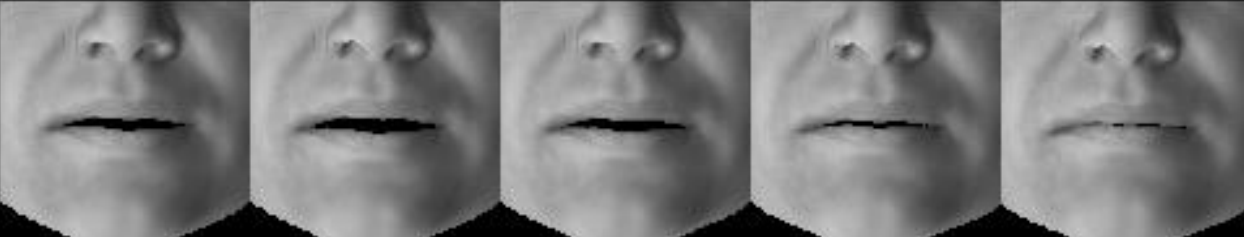}
    \caption{Example of the cropped interesting region in the 3D rendered face sequences to calculate the lip-reading loss.}
    \label{fig:mouth}
\end{figure}

Besides, we use pretrained lip-reading models proposed in \cite{ma2022training} to calculate lip reading loss $\mathcal{L}_{read}$ inspired by \cite{spectre}. We use the pretrained video-based lip-reading model where the input is a sequence~(5 frames in our case) of the cropped interesting region around the mouth~(as shown in Fig~\ref{fig:mouth}) and the target is a series of the character sequence. So we employ a differentiable 3D face render~\cite{deng2019accurate} in ExpNet to render the images through the generated expression coefficients, then, we crop the mouth area of the rendered images using the bounding box of the mouth landmarks, obtaining the logit of the character sequences $\mathbf{C_p}$. As for the supervision, we generate the logit of the character sequences $\mathbf{C_{gt}}$ from the ground truth audio using the audio-driven lip-reading model. Thus, our goal is to minimize the difference between $\mathbf{C_p}$ and $\mathbf{C_{gt}}$. In other words,  
\begin{equation}
    \mathcal{L}_{read}= \text{CrossEntrory}(\mathbf{C_{gt}}, \mathbf{C_p})
\end{equation}
Overall, the final loss of ExpNet is given by :
\begin{equation}
    \mathcal{L}_{exp}=\lambda_{distill} \mathcal{L}_{distill}+\lambda_{read} \mathcal{L}_{read} + \lambda_{lks} \mathcal{L}_{lks}
\end{equation}
Where $\lambda_{distill}$, $\lambda_{read}$, $\lambda_{lks}$ are set to $2$, $0.01$, and $0.01$, respectively.

\paragraph{PoseVAE} We first calculate the reconstruction loss by applying Mean-Squared loss between the generated $\Delta \rho^\prime_{\{1 \ldots T\}}$ and the original $\Delta \rho_{\{1 \ldots T\}}$: 
\begin{equation}
\mathcal{L}_{MSE} = \frac{1}{T} \sum_{t=1}^{T}\left(\Delta \rho^{\prime}_t - \Delta \rho_t \right)^{2}
\end{equation}
Meanwhile, we encourage the similarity of the latent space distribution  and the Gaussian distribution with the mean vector ${\mu}$ and covariance matrix ${\sum}$. So we define $\mathcal{L}_{KL}$ as the Kullback–Leibler~(KL) divergence between the latent space distribution and the Gaussian distribution.
We also employ a discriminator $D$ based on the PatchGAN~\cite{pix2pix} to perform 1D convolution on the head motion sequence as Speech2Gesture~\cite{speech2gesture}. We define the adversarial loss $\mathcal{L}_{GAN}$:
\begin{equation}
\mathcal{L}_{GAN} = \arg \min _{G} \max _{D}\left(G, D\right)
\end{equation}
Where $G$ is proposed PoseVAE. The total loss of PoseVAE can be summarized as follows. 
\begin{equation}
\mathcal{L}_{pose} = \lambda_{MSE} \mathcal{L}_{MSE}+\lambda_{KL} \mathcal{L}_{KL} + \lambda_{GAN} \mathcal{L}_{GAN}
\end{equation}
where $\lambda_{MSE}$, $\lambda_{KL}$ and $\mathcal{L}_{GAN}$ are set to 1, 1, and 0.7, respectively.

\paragraph{FaceRender}
We add a MappingNet to map the explicit 3DMM coefficients to the space of the face-vid2vid~\cite{facevid2vid}, to training, apart from the loss functions used in face-vi2vid~\cite{facevid2vid}, we add $L_1$ regularization on the domain of unsupervised keypoints:
\begin{equation}
\mathcal{L}_{1} = \frac{1}{N} \sum_{n=1}^{N}||K^{\prime}_n - K_n ||_{1},
\end{equation}
where $K^{\prime}_n$ and $K_n $ are the $n$-th keypoint generated by our MappingNet and the motion generator of the original face-vid2vid, respectively. The weight of $\mathcal{L}_{1}$ is set to 20, and the weights of the other loss functions keep the same as in face-vid2vid and they are calculated on the final generated image. Please refer to face-vid2vid~\cite{facevid2vid}
for more details.

\subsection{More Details about the Alignment Coefficients.}
\label{supp:align}
In the main paper, we show the effect of the alignment coefficients in Fig~\ref{fig:render_ablation}. Here, we give more details about the alignment coefficients. Generally, the alignment coefficients are the transformation parameters~(translation and scaling) to transform and crop the arbitrary video to the aligned face video for deep 3D face reconstruction~\cite{deng2019accurate}. The implicit modulation of PIRenderer~\cite{ren2021pirenderer} contains 73 dimensional motion coefficients, including the expression~(64), head pose~(6) and the alignment coefficients~(3). Since their method focuses on video driving animation, the alignment coefficients are known in testing. However, it is hard to \textit{learn} from the audio since there is no relationship between the alignment and the audio. We also try to learn and use the alignment coefficients of the first frame in our method~(as shown in Fig.~\ref{fig:render_ablation} and the supp. video), however, the produced head motion is aligned and unnatural. We discard it to obtain more natural video results. Thus, our motion coefficients~(70) only contains the expression~(64) and head pose~(6).

\section{Supplementary Video}
We provide a supplementary video to include all the video results of our method and other related methods as comparisons, the ablation study of each component, and more results of our method in different languages.

     \fi

\end{document}